%% file: ms.tex
\documentclass{article}

     \usepackage[preprint]{neurips_2022}

\usepackage[utf8]{inputenc} 
\usepackage[T1]{fontenc}    
\usepackage{hyperref}       
\usepackage{url}            
\usepackage{booktabs}       
\usepackage{amsfonts}       
\usepackage{nicefrac}       
\usepackage{microtype}      
\usepackage{xcolor}         

\usepackage{graphicx}
\usepackage{amsmath}
\usepackage{amssymb}
\usepackage{mathtools}
\usepackage{scalerel}
\usepackage{amsthm}
\usepackage{algorithm}
\usepackage{algcompatible}
\usepackage{bbm}
\usepackage{enumitem}
\usepackage{multirow}
\usepackage[capitalize,noabbrev]{cleveref}
\usepackage[textsize=tiny]{todonotes}

\newlist{Properties}{enumerate}{1}
\setlist[Properties]{label=Property \arabic*., font=\textbf, itemindent=*}

\newtheorem{proposition}{Proposition}
\newtheorem{lemma}{Lemma}
\newcommand{\mc}[1]{\mathcal{#1}}
\newcommand{\mb}[1]{\mathbb{#1}}
\newcommand{\ts}{\textsuperscript}
\newcommand{\ie}{{\em i.e.~}}
\newcommand{\Prob}{\mathbb{P}}
\newcommand{\argmin}{\mathop{\mathrm{arg\,min}}}
\newcommand{\norm}[1]{\left \lVert #1 \right \rVert}
\newcommand{\hst}{\hspace{0.2cm}}

\newcommand{\0}{\mathbf {0}}
\def\bbR{\mathbb{R}}
\title{A Confidence Machine for Sparse High-Order Interaction Model}
\author{%
  Diptesh~Das \thanks{corresponding authors}\\
  Nagoya University, Japan\\
  \texttt{diptesh.das@mae.nagoya-u.ac.jp} \\
  \And
 Eugene~Ndiaye\\
  Georgia Institute of Technology, USA \\
  \texttt{endiaye3@gatech.edu} \\
  \AND
  Ichiro~Takeuchi \footnotemark[1]\\
  Nagoya University / RIKEN, Japan\\
  \texttt{ichiro.takeuchi@mae.nagoya-u.ac.jp} \\
}

\begin{document}

\maketitle

\input{abstract}
\input{sec1}
\input{sec2}
\input{sec3}
\input{sec4}
\input{conclusion}
\input{appendix-A}
\input{appendix-B}
\input{appendix-C}
\begin{ack}
\input{acknowledgement}
\end{ack}

\bibliographystyle{ACM-Reference-Format}
\bibliography{ms}

\end{document}

%% file: abstract.tex
\begin{abstract}
In predictive modeling for high-stake decision-making, predictors must be not only accurate but also reliable. Conformal prediction (CP) is a promising approach for obtaining the confidence of prediction results with fewer theoretical assumptions. To obtain the confidence set by so-called full-CP, we need to refit the predictor for all possible values of prediction results, which is only possible for simple predictors. For complex predictors such as random forests (RFs) or neural networks (NNs), split-CP is often employed where the data is split into two parts: one part for fitting and another to compute the confidence set. Unfortunately, because of the reduced sample size, split-CP is inferior to full-CP both in fitting as well as confidence set computation. In this paper, we develop a full-CP of sparse high-order interaction model (SHIM), which is sufficiently flexible as it can take into account high-order interactions among variables. We resolve the computational challenge for full-CP of SHIM by introducing a novel approach called homotopy mining. Through numerical experiments, we demonstrate that SHIM is as accurate as complex predictors such as RF and NN and enjoys the superior statistical power of full-CP. 
\end{abstract}

%% file: sec1.tex
\section{Introduction}\label{sec:introduction}
The uncertainty in data-driven analysis is a major concern, particularly in risk-sensitive automated decision-making problems (for example, in medical diagnosis and criminal justice). Several strategies exist to quantify the uncertainty of a point estimators. For example, the Bayesian approach can provide a strong confidence bound, but requires the assumption on prior distribution. 
%
%
The PAC analysis is another approach that provides bounds on the probability of error. %
As another direction, selective inference has been studied for quantifying the uncertainty of data-driven knowledge \cite{das2021fast,le2021parametric,duy2020computing}.
The conformal prediction (CP) is one such uncertainty quantification method that is very generic, and it is applicable to almost any point estimators \cite{vovk2005algorithmic,shafer2008tutorial}. 
The CP method has recently gained significant attention as it can provide valid finite sample statistical coverage guarantee at any nominal level as long as data are independently and identically distributed (i.i.d.). The coverage guarantee provided is valid even when the model is misspecified. In this paper we are interested in the \emph{full-CP} where the full data is used to compute the CP set. An alternative to this approach is the \emph{split-CP} in which the data is split into two parts: one part is used for fitting and the remaining part is used for computing the CP set.
The essential idea of the full-CP framework can be stated as follows: Given a training set \(\mc{D}_n = \{(x_1, y_1), \ldots, (x_n, y_n)\}\) and a test instance $x_{n+1}$ which are both i.i.d., the goal of CP is to construct a $100(1-\alpha)\%$ confidence set that contains the unobserved $y_{n+1}$. Here, $\alpha \in [0,1]$ represents the confidence level. In other words, we are interested in all possible completions of the augmented dataset in the form of \(\mc{D}_n \cup (x_{n+1}, \tau)\), such that $y_{n+1}=\tau$ is \textit{typical} for the known data  \(\{\mc{D}_n, x_{n+1}\}\). This typicality is measured by a typicalness function $\pi(\cdot)$, which is also called the \emph{$p$-value} in analogy to the classical hypothesis testing. In other words, the CP set for $x_{n+1}$ is defined as the set of all $\tau$ for which the null hypothesis $H_0: y_{n+1}=\tau$ is not rejected against the alternative hypothesis  $H_1: y_{n+1}\neq\tau$. \\

\textbf{Related work:} Since its inception, there have been several extensions and applications of the CP framework in diverse directions. Examples include the choice of conformity score and statistical efficiency \cite{lei2013distribution,lei2014distribution}, high-dimensional regression \cite{lei2018distribution}, classification \cite{lei2014classification,sadinle2019least}, active learning \cite{ho2008query}, time series \cite{xu2021conformal}, few-shot learning \cite{fisch2021few}, text and speech completion \cite{dey2021conformal}, image classification \cite{angelopoulos2020uncertainty}, outlier detection \cite{laxhammar2015inductive,bates2021testing}. 
%
Recently, some approximate computation of the conformal set was proposed for regression in \cite{ndiaye2019computing,ndiaye2021root} and in classification, significant advancement has been made in \cite{cherubin2021exact} to compute exact full-CP.
\begin{figure}[t]
  \centering
  \includegraphics[width=\linewidth]{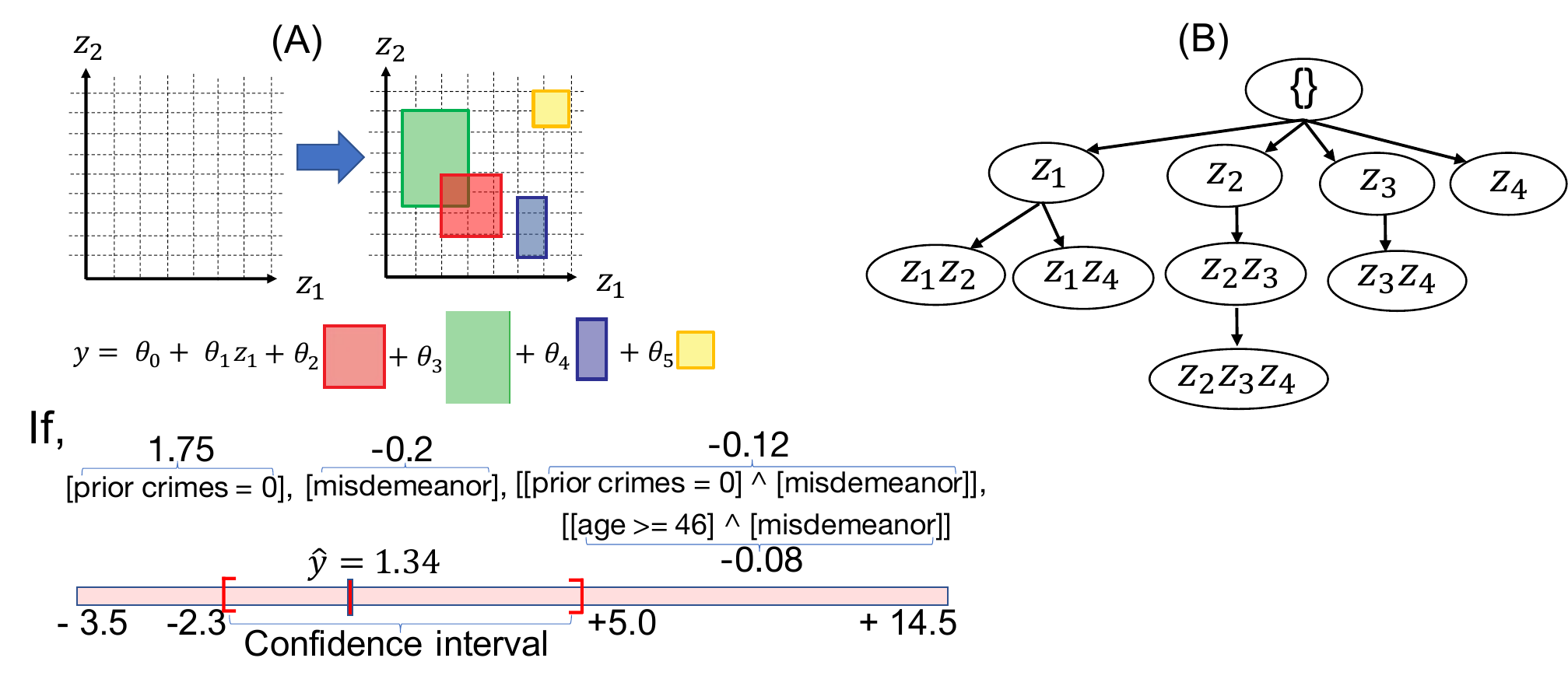}
  \caption{(A) An intuitive illustration of a SHIM solution space, which is a hyper rectangle (\ie, generalization of a rectangle for higher dimensions that essentially represents a cartesian product of intervals). The decision sets, the predicted response $(\hat{y})$ and the associated confidence interval for a randomly chosen example of data from the ProPublica two-year recidivism criminal justice (COMPAS) dataset \cite{larson2016we} are shown. (B) A tree of patterns has been constructed by exploiting the hierarchical structure of high-order interaction features (Note: not all patterns appear due to pruning).\looseness=-1
}
\label{fig1}
 \end{figure}
Despite its attractive properties, the application of exact, full-CP in many practical problems  remains an open problem owing to its high computational cost. 
By definition, the full-CP framework requires the refitting of model by augmenting the data for every possible candidate $\forall \tau \in \mb{R}$ of the unobserved $y_{n+1}$. 
%
%
In a regression setting, one needs to refit the model an infinite number of times for all possible candidates on the real line $(\forall \tau \in \mb{R})$, and check the conformity by computing the $p$-values $\pi(\tau)$. \looseness=-1

Hence, an efficient computation of the full-CP set is possible only for a handful of simple models (e.g., ordinary LS regression, ridge regression) in which the solution is explicitly represented as a function of $\tau$.
This enables a closed-form derivation of full-CP set and avoids an exhaustive search over the real line \cite{nouretdinov2001ridge}. 
Recently, \cite{lei2019fast} proposed a homotopy method to efficiently compute the full-CP set of the LASSO. The homotopy method exploits the piece-wise linearity of the LASSO solutions and avoids the computational burden of all possible candidates of $y_{n+1}$ (see section~\ref{proposed_method} for details).
%
%
%
%
%
%
Unfortunately, such a structure does not exist for most of the models, and the application of full-CP is still an open question for complex models that can represent complex nonlinearity such as a random forest (RF), neural network (NN) etc; for which split-CP has been the only possible choice.\looseness=-1

%
%
%
%
In CP, although the coverage property is satisfied even when the model is misspecified, it is better to use sufficiently complex models for complex data because this enables us to obtain a more compact confidence set (\ie, shorter confidence intervals). Furthermore, it is important to note that the confidence set obtained by a full-CP is more compact than that obtained by a split-CP because only a part of instances in the available dataset is used for constructing the confidence set in split-CP. This means that it is valuable to construct a full-CP algorithm for sufficiently complex models.
%
%
%
%
%

In this paper, we considered the \emph{sparse high-order interaction model (SHIM)} which can represent complex nonlinear relationship, and proposed a \emph{homotopy-mining} method to efficiently compute the \emph{full-CP} set. We call the resulting machine as \emph{SHIM confidence machine}. SHIM is formulated as a weighted sum of conjunction rules that are highly interpretable as well as accurate decision sets \cite{das2019interpretable,lakkaraju2016interpretable}. SHIM can capture the combinatorial interactions of multiple factors, which can prove to be beneficial for deciphering complex data. A conjunction rule of a SHIM looks like \(I(-1.5 \leq x_1 \leq 2.3)\wedge I(x_3 \geq 20.0)\wedge I(x_5 \leq 7.5) \), where $I(\cdot)$ refers to the indicator function. An intuitive illustration of a SHIM is shown in \figurename~\ref{fig1}A. \phantom{haha}

%
%
\vspace{0.2cm}
\textbf{Contribution:}
Our main contribution in this paper is to develop an efficient algorithm to conduct \emph{exact full-CP} for SHIM which can capture complex structures in the data by considering high-order interaction features.
The exact full-CP for complex black box models such as RF and NN are intractable and an efficient computational method does not exist. For such black box models only split-CP has been possible.
As of today, an efficient exact full-CP is possible only for simple regression models such as LASSO, ridge regression, and ordinary least square regression. The proposed method adds SHIM to that list. To the best of our knowledge, this is also the first attempt to construct a conformal prediction set in the context of pattern mining model which is fitted with branch and bound approach.
We also extended our framework to the elastic net and provided an algorithm to compute the exact full-CP set efficiently.
%
%
%

The proposed method enables us to obtain a more compact confidence set (much better than linear and comparable to non linear models) by the full-CP for sufficiently flexible and complex SHIM.
A SHIM uses higher-order interaction features and hence, the full-CP is computationally challenging; but we overcome this difficulty by introducing a method called \emph{homotopy mining} which exploits the best of both homotopy and (pattern) mining methods. The computation of exact full-CP for SHIM by homotopy mining can be interpreted as an extension of LASSO's exact full-CP in \cite{lei2019fast}.
The use of black box or explainable machine learning models in high-stakes decision-making such as in healthcare, criminal justice, and other domains is highly criticized in the literature \cite{angelino2018learning,rudin2019stop}. We believe that the exact full-CP of a SHIM has significant importance in practice where accuracy, statistical reliability, and interpretability are important. If a practitioner chooses SHIM as a regressor, how to compute the corresponding exact full-CP set? The proposed method provides a computationally efficient solution to this. 

\textbf{Code:} The source code is available at
\url{https://github.com/DipteshDas/CP-SHIM}.\looseness=-1 
\vspace{0.2cm}

\textbf{Notation:}
In our notation, $y\in \mathbb{R}^n$ represents the response vector of $n$ instances, $y(\tau) \in \mathbb{R}^{n+1}$ is the augmented response vector constructed by augmenting the possible response value ($\tau \in \mb{R}$) of the $(n+1)^{th}$ instance with $y$. Later we defined $y(\tau)$ as a function of the variable $\tau$ as the vector $y(\tau)$ changes for different possible response value ($\tau$) of the $(n+1)^{th}$ instance. The $y_i\in \mathbb{R}, \forall i \in [n+1]$, represents the scalar response of the $i^{th}$ instance, $X \in \mb{R}^{(n+1) \times p}$ is the design matrix of all $(n+1)$ instances, each having $p$ features, $x_\ell \in \mb{R}^{n+1}$ is the column vector corresponding to some $\ell \in [p]$, where $[p] = \{1, \ldots, p\}$, $X_{\mc{A}_\tau} \in \mb{R}^{(n+1) \times |\mc{A}_\tau|} \subseteq X$ is a smaller design matrix in which the columns are restricted to the elements of some subset $\mc{A}_\tau \subseteq [p]$.  \looseness=-1
%
%
%

%% file: sec2.tex
\section{Problem Statement}
Consider a regression problem with a response vector $y \in \mb{R}^n$ and 
$m$ original covariate vectors $z_1, \ldots, z_m,$ where \(z_{\ell} \in \mb{R}^n\) and $\ell \in [m] = \{1, ..., m\}$. A high-order interaction model up to the $d^{\rm th}$ order is then written as follows:
\begin{equation}\label{eq:shim_model}
\begin{split}
\hspace{-2.5mm} y = 
\sum_{\ell_1 \in [m]} \theta_{\ell_1} z_{\ell_1} 
+ \sum_{ \substack{ (\ell_1, \ell_2)  \in [m] \times [m]  \\ \ell_1 \neq \ell_2}} \hspace{-2.5mm} \theta_{\ell_1, \ell_2} z_{\ell_1} z_{\ell_2}
+  \cdots
+ \sum_{\substack{(\ell_1, ..., \ell_d)  \in [m]^d \\ \ell_1 \neq ... \neq \ell_d}} \hspace{-2.5mm} \theta_{\ell_1, \ldots, \ell_d} z_{\ell_1} \cdots  z_{\ell_d},
\end{split}
\end{equation}
where $z_{\ell_1} \cdots  z_{\ell_d}$ is the element-wise product and scalar $\theta$ represents the coefficient.
In this study, we mainly consider each element of the original covariate vector $z_{\ell} \in \{0, 1\}^n$ for $\ell \in [m]$. However, our model is equally applicable to covariate vectors defined in the domain $[0, 1]^n$.
To simplify the notation, it is convenient to write the high-order interaction model in (\ref{eq:shim_model}) using the following matrix of concatenated vectors of all high-order interactions:
\[
X=[\underbrace{z_1, \ldots, z_m}_{1\text{\ts{st} order}}, 
\cdots, \underbrace{z_1 \ldots z_d, \ldots, z_{m-d+1} \ldots z_m}_{d\text{\ts{th} order}}] \in \mb{R}^{n \times p},
\]
where \(p := \sum_{\kappa=1}^d{m \choose \kappa}\).
Similarly, the coefficient vector associated with all possible high-order interaction terms can be written as follows:
\[
\beta:= [ \underbrace{ \theta_1, \ldots, \theta_m}_{1\text{\ts{st} order}},
\cdots, \underbrace{\theta_{1, \ldots ,d},  \ldots, \theta_{m-d+1, \ldots, m}}_{d\text{\ts{th} order}}
]^\top \in \mb{R}^p.
\]
The high-order interaction model (\ref{eq:shim_model}) is then simply written as a linear model
$
y = X \beta.
$
Unfortunately, $p$ can be prohibitively large unless both $m$ and $d$ are fairly small.
In SHIM, we consider a sparse estimation of a high-order interaction model.
An example of SHIM is as follows:
\begin{equation*}\label{eq:shim_eg}
 y = \theta_3z_{3} +  \theta_5z_{5} +  \theta_{2,6}z_{2}z_{6} +  \theta_{1,2,5,9}z_{1}z_{2}z_{5}z_{9}.
\end{equation*}
Before delving into our proposed method, we briefly overview the conformal prediction framework.
\subsection{Conformal prediction}
A mere point estimation is insufficient for risk-sensitive automated decision-making problems \cite{rudin2019stop,angelino2018learning,das2019interpretable}, such as in medical diagnosis and criminal justice. In such high-stake decision-making problems, if the estimators are equipped with the associated confidence information, the decision maker will be more informed and sufficiently confident to make a prudent decision when the stakes are high.   

\textbf{Full-CP:}
Given a labelled dataset ${\mc{D}}_n$ and a new observation $x_{n+1}$, the goal of full-CP framework is to construct a set of likely values $\mc{C}(x_{n+1})$ of unobserved $y_{n+1}$ with a valid statistical coverage guarantee \cite{vovk2005algorithmic,shafer2008tutorial}, \ie,
\begin{equation}\label{eqn:CP_guarantee}
    \Prob(y_{n+1} \in \mc{C}(x_{n+1})) \geq 1 - \alpha,
\end{equation}
where \(\alpha \in [0, 1]\) determines the level of confidence. 
If we define a prediction function  $\mu(\cdot)$ that maps the input $X$ to the output $y$, then the essential idea of constructing a full-CP set is to fit a model $\mu_{\tau}(\cdot)$ with the augmented data $\mc{D}_n \cup (x_{n+1},\tau)$ for every possible candidate $\tau \in \mb{R}$ and compare the prediction error of each instances.
More precisely, let \( y(\tau)  = [y_1, \ldots, y_n, \tau]^\top \in \mb{R}^{n+1} \) be a vector augmented with $\tau$. We define a score function that measures how well the model can predict each output variables; with the constraint that it should not depend on the order of the data instances \cite{lei2019fast}. One such conformity score function generally used in linear regression setting is the (coordinate-wise) absolute residual \ie,
 $$S(\tau) = | y(\tau) - \mu_{\scaleto{\tau\mathstrut}{6pt}}(X)|,$$
 where we stack the input vectors in the design matrix $X = [x_1, \ldots, x_{n+1}]^\top$ in $\mb{R}^{(n+1) \times p}$.
The conformity function can now be defined as:
$$ \pi (\tau) = 1 - \frac{1}{n+1}\sum_{i=1}^{n+1} \mathbbm{1}_{S_i(\tau) \leq S_{n+1}(\tau)},$$
which evaluates how the prediction of the candidate $\tau$ using the fitted model $\mu_{\tau}(x_{n+1})$ is ranked compared to the prediction of the previously observed data $y_{i}$ using $\mu_{\tau}(x_{i})$. The conformal set is defined as:
\begin{equation}\label{eqn:CP2}
    \mc{C}(x_{n+1}) = \{\tau:  \pi (\tau) \geq \alpha\},
\end{equation}
which is merely the collection of candidates whose conformity is large enough. In \cite{vovk2005algorithmic}, the author showed that the defined conformal set $\mc{C}(x_{n+1})$ satisfies the coverage guarantee (\ref{eqn:CP_guarantee}) as long as the data are i.i.d.

\textbf{Split-CP:} In split-CP \cite{papadopoulos2002inductive}, the data set \(\mc{D}_n\) is split into two parts, defined as the training set \(\mc{D}_{\text{tr}}=\{(x_1, y_1), \ldots, (x_m, y_m)\}\) and the calibration set \(\mc{D}_{\text{cal}}=\{(x_{m+1}, y_{m+1}), \ldots, (x_n, y_n)\}\), with \(m < n\). Then the model ($\mu^{\text{tr}}(\cdot)$) is fit with the training set $\mc{D}_{\text{tr}}$ only once, and the $p$-values, denoted as $\pi_{\text{split}}(\cdot)$, are determined using the calibration set $\mc{D}_{\text{cal}}$ as follows.
%
%
%
\begin{equation*}\label{eqn:pi_z_split}
    \pi_{\text{split}} (\tau) = 1 - \frac{1}{n-m}\sum_{i=m+1}^{n} \mathbbm{1}_{S^{\text{cal}}_i(\tau) \leq S_{n+1}(\tau)},
\end{equation*}
where \(S^{\text{cal}}_i(\tau) = |y_i - \mu^{\text{tr}}(x_i)|, \forall i \in [m+1, n]\).
%
%
%
Therefore, the split-CP can be defined as follows:
\begin{align*}\label{eqn:CP_split}
    \mc{C}_{\text{split}}(x_{n+1}) = \{\tau:\pi_{\text{split}} (\tau) \geq \alpha\},
\end{align*}
where \(Q^{\text{cal}}_{1-\alpha}\) is the \((1-\alpha)\) quantile of the calibration scores \(S^{\text{cal}}_i, \forall i \in [m+1, n]\). When the conformity score is defined as the absolute residual, then the split-CP set can be conveniently written as \(\mc{C}_{\text{split}}(x_{n+1}) = [\mu^{\text{tr}}(x_{n+1}) \pm Q^{\text{cal}}_{1-\alpha}] \).

Although split-CP enjoys the computational efficiency of single model fitting, it suffers from the poor statistical efficiency owing to the smaller sample size both in the model fitting and calibration phases. 



On the other hand, to compute the exact full-CP set for a regression problem where \(y \in \mb{R}\), one needs to refit the regression model $\mu_\tau(\cdot)$ and evaluate $\pi(\tau)$ for an infinite number of candidates $\tau$. Hence, efficient computation of the exact full-CP set is possible only for a handful of predictors (e.g., ridge regression) where it is possible to have a closed form expression of $\mu_{\tau}(\cdot)$. 
However, such structure does not exist for most of the machine learning models and computing the exact full-CP set remains an open question for many models. Recently, in \cite{lei2019fast} the author proposed a homotopy method to efficiently compute the exact full-CP set of the LASSO which was not trivial. 
They considered the response $y$ as a function of a scalar and derived a homotopy method to construct the exact LASSO solution path as function of that scalar. After constructing the exact path, the solutions at the transition points of the path were used to efficiently compute the exact full-CP set of the LASSO. In this paper we extend that framework for the SHIM.

%

%
%

%% file: sec3.tex
\section{Proposed Method}\label{proposed_method}
%
%
%
We propose a \emph{homotopy-mining} method to compute the exact full-CP set of a SHIM. 
The homotopy method refers to an optimization framework for solving a sequence of parameterized optimization problems.  
The basic idea of our method is to consider the following optimization problem: 
\begin{equation}\label{obj:primal_lasso}
    \beta(\tau) \in \argmin_{\beta \in \mb{R}^p} \frac{1}{2}\norm{y(\tau) - X\beta}_2^2  + \lambda \norm{\beta}_1.
\end{equation}
%
%
At optima, it holds
$$ X^\top \big(X\beta(\tau) - y(\tau)\big) + \lambda s(\tau) = 0,$$
 where for any $\ell$ in $[p]$,
\begin{equation*}\label{eq:opt_condn_shim_homotopy}
    s_{\ell}(\tau) \in \begin{cases}
        \{ -1, +1 \} &\text{if} \hst \beta_\ell(\tau) \neq 0,\\
        [-1, +1]  &\text{if} \hst   \beta_\ell(\tau) = 0,
\end{cases}
\end{equation*}
%
%
Let us define the active set
\begin{align}\label{eq:active_set}
\mc{A}_\tau = \left\{\ell \in [p]:\; \big\lvert x_{\ell}^{\top} w(\tau) \big\rvert = \lambda \right\}
\end{align}
where 
\begin{align}
  w(\tau)  = y(\tau) - X\beta(\tau)
\end{align} 
is the residual~\cite{efron2004least}.
%
%
In \cite{lei2019fast}, the author showed that for a fixed $\lambda$, the exact solution path of LASSO with a variable response $y(\tau)$, characterized by $\tau$ can be shown to be piece-wise linear as stated in Proposition \ref{prop:homotopy}.
\begin{proposition}\label{prop:homotopy}
If $\beta(\tau)$ have the same sign between two points $\tau_1$ and $\tau_2$ \ie,
%
$\mathrm{sign}(\beta(\tau_1)) = \mathrm{sign}(\beta(\tau_2)) = \mathrm{sign}(\beta(\tau))$ for any $\tau \in [\tau_1, \tau_2)$,
then $\mc{A}_{\tau} = \mc{A}_{\tau_1}$. Furthermore, assuming that \(X^{\top}_{\mc{A}_{\tau}}X_{\mc{A}_{\tau}}\) is invertible, we have the linear relations
\begin{align*}
\beta_{\mc{A}_{\tau}} (\tau_2) &= \beta_{\mc{A}_{\tau}} (\tau_1) + (\tau_2 - \tau_1) \times \nu_{\mc{A}_\tau}(\tau),\\
s_{\mc{A}^c_{\tau}} (\tau_2)  &= s_{\mc{A}^c_{\tau}} (\tau_1) +  (\tau_2 - \tau_1) \times \gamma_{\mc{A}^c_\tau}(\tau) / \lambda,
\end{align*}
where the direction vectors are defined as
\begin{align*}
\nu_{\mc{A}_\tau}(\tau) &= (X^{\top}_{\mc{A}_{\tau}}X_{\mc{A}_{\tau}})^{-1}x_{n+1,\mc{A}_{\tau}},\\
\gamma_{\mc{A}^c_\tau}(\tau) &= x_{n+1, \mc{A}_{\tau}^c} - (X^{\top}_{\mc{A}_{\tau}^c}X_{\mc{A}_{\tau}})\nu_{\mc{A}_\tau}(\tau).
\end{align*}
For simplicity, we will denote the step size $\Delta = \tau_2 - \tau_1 > 0$. 
%

%
%
\end{proposition}
%
%
%

%
%
We call the mapping \(\tau \mapsto \beta(\tau)\) the \emph{$\tau$-path} and the number of linear pieces of this $\tau$-path is upper bounded by the number of all possible signs $3^p$. Note that the possible values of $\mathrm{sign}(\beta)$ are $-1, 0$ and $+1$.
%
This $\tau$-path can be computed exactly using homotopy method \cite{efron2004least, rosset2007piecewise, mairal2012complexity} that sequencially tracks and updates the sign and active set of the optimal solution by exploiting its linearity between each two consecutive transition points of direction ($\nu_{\mc{A}_\tau}(\tau),\gamma_{\mc{A}^c_\tau}(\tau)$) changes. 
At every consecutive step, represented by $\tau_t, \tau_{t+1}$, where $t$ is an index of the transition points (kinks) of the $\tau$-path, either of the following two events occurs:\looseness=-1
\vspace{0.2cm}
\begin{itemize}
 \item a zero variable becomes non-zero, that is, 
 $$ \exists \ell \in \mc{A}^c_{\tau_t} \text{ s.t. } \big\lvert x_{\ell}^{\top} w(\tau_{t+1}) \big\rvert = \lambda \hst \text{or,}$$
\item a non-zero variable becomes zero, that is,
$$\exists \ell \in \mc{A}_{\tau_t} \text{ s.t. } \beta_\ell(\tau_t) \neq 0 \text{, but } \beta_\ell(\tau_{t+1}) = 0.$$
\end{itemize}
Overall, the next change in the active set (or change in direction vectors) occurs at $\tau_{t + 1} = \tau_t + \Delta_{\ell^\ast}$, such that 
\begin{equation}\label{eq:step_length}
\Delta_{\ell^\ast} = \min(\Delta_1(\ell_{1}^{\ast}), \Delta_2(\ell_{2}^{\ast})),
\end{equation}
%
%
%
%
%
where
\begin{align*}
  &\ell_{1}^{\ast} = \argmin_{\ell \in \mc{A}_{\tau_t}} \Delta_1(\ell),\\
  &\ell_{2}^{\ast} = \argmin_{\ell \in \mathcal{A}^c_{\tau_t}} \Delta_2(\ell), \text{  and}
  \end{align*}
\begin{align*}
  \Delta_1(\ell) &= \left(- \frac{\beta_\ell(\tau_t)}{\nu_\ell(\tau_t)}\right)_{++}, \\
\Delta_2(\ell) &= \left( \lambda \frac{ \text{sign}(\gamma_\ell (\tau_t)) - x_{\ell}^{\top}w(\tau_{t})}{\gamma_\ell(\tau_t)} \right)_{++}.
\end{align*}
Here, we use the convention that for any $a \in \mb{R}, (a)_{++} = a$ if $a > 0$ and $\infty$ otherwise. However, naively (minimization over all possible interaction terms) determining the step size of inclusion $(\Delta_2({\ell_2^\ast}))$ will be intractable for the SHIM type problem. Because in SHIM the search space grows exponentially due to the combinatorial effect of high-order interaction terms. 
%
%
Therefore, both the fitting and constructing the full-CP set of a SHIM are non-trivial because unless both $m$ and $d$ are very small, a high-order interaction model will have a significantly large number of parameters to be considered.
Several algorithms for fitting a sparse high-order interaction model have been proposed in the literature \cite{ERP_Tsuda,saigo2009gboost,nakagawa2016safe}.
A common approach adopted in these existing works is to exploit the hierarchical structure of high-order interaction features.
In other words, a tree structure as in Fig.~\ref{fig1}B is considered and a branch-and-bound strategy is employed in order to avoid handling all the exponentially increasing number of high-order interaction features. Hence, we need efficient computational methods to make the computation practically feasible. 
In the following section, we present an efficient tree pruning strategy that considers the tree structure of the interaction terms (or patterns). Here, each node of the tree represents an interaction term. The basic idea of tree pruning is that we construct a tree of interaction terms in a ``progressive manner". 
That is we keep track of the current minimum step size of inclusion up to the construction of $\ell^{th}$ pattern as we construct the tree progressively, and prune a large part of the tree if some bound condition fails (Lemma~\ref{lemma:lemma_1}). 
%

\subsection{Tree pruning ($\tau$-path)}\label{tree_pruning}
%
%
%
%
%
\textbf{Definition of Tree:}
A tree is constructed in such a way that for any pair of nodes (\(\ell, \ell^\prime)\), where $\ell$ is the ancestor of $\ell^\prime$, \ie, \(\ell \subset \ell^\prime\), the following conditions are satisfied
\begin{align*}
&x_{i \ell^\prime} = 1 \implies x_{i\ell } = 1,
     &x_{i\ell } = 0 \implies x_{i \ell^\prime} = 0, \forall i \in [n+1].
\end{align*}
The basic idea of our tree pruning condition is stated below. The equicorrelation condition for any active feature \(k  \in \mathcal{A}_{\tau}\) at a fixed $\lambda$ can be written as
$\big\lvert x_{k}^{\top} w(\tau) \big\rvert = \lambda$ (see definition of $\mc{A}_{\tau}$ in (\ref{eq:active_set})).
%
%

Therefore, at $\tau = \tau_{t + 1}$ such that $\tau_{t+1} = \tau_t + \Delta_2(\ell)$ any non-active feature \(\ell \in \mathcal{A}^c_{\tau_{t} } \) becomes active  if
\begin{equation}\label{eqn:inclusion_condition2}
    \big\lvert x_{\ell}^{\top} w(\tau_{t+1})\lvert =
    \big\lvert x_{k}^{\top} w(\tau_{t+1}) \big\rvert .
\end{equation}
Now, using the triangular inequality one can show that (\ref{eqn:inclusion_condition2}) will not have any solution if
\begin{align}\label{cond:pruning_tau1}
\nonumber
\lvert \rho_{\ell}(\tau_t) \lvert + &\Delta_2(\ell) (\lvert \eta_{\ell}(\tau_t) \lvert + x_{n+1,\ell}) < \\
&\lvert \rho_k (\tau_t) \lvert   - \Delta_2(\ell)(\lvert\eta_k(\tau_t) \lvert +  x_{n+1,k}),
\end{align}
where,
\begin{align*}
    \rho_{\ell} &= x_{\ell}^{\top}w(\tau_t)  \text{ and }
    \eta_{\ell} = x_{\ell}^{\top} v(\tau_t) \quad \forall \ell \in \mathcal{A}^c_{\tau_t},\\
    \rho_k &= x_{k}^{\top}w(\tau_t) \text{ and }
    \eta_k = x_k^{\top} v(\tau_t) \quad \forall k \in \mathcal{A}_{\tau_t},\\
    v(\tau_t) &= X_{\mc{A}_{\tau_t}}\nu_{\mc{A}_{\tau_t}}(\tau_t).
\end{align*}
Therefore, (\ref{cond:pruning_tau1}) can be used to derive the pruning condition of the $\tau$-path which is formally stated in Lemma \ref{lemma:lemma_1}.\\
%

Similar idea has been used in the context of graph mining  \cite{ERP_Tsuda} and selective inference of SHIM \cite{das2021fast}. Note that \cite{ERP_Tsuda} provided a pruning condition for the exact regularization of graph data ($\lambda$-path) and \cite{das2021fast} provided a pruning condition in the context of selective inference to characterize the conditional distribution of the test statistics. However, in our case we adapted the similar idea to compute the exact full-CP set of SHIM. 
%
%
%
\begin{lemma}\label{lemma:lemma_1}
For any given node $\ell$, if $\Delta_2(\ell_2^\dagger)$ is the current minimum step size, that is,
$$
\ell_{2}^{\dagger} = \argmin_{j \in \{1, 2, \ldots, \ell\} \cap \mc{A}_\tau^c} \Delta_2(j)
$$
%
then
$\forall \ell^\prime \supset \ell, \Delta_2({\ell^\prime}) \geq \Delta_2({\ell_2^\dagger})$ if 
\begin{align} \label{eq:lemma_1_eq}
\nonumber
b_{\ell} (w(\tau_t)) + &\Delta_2({\ell_2^\dagger}) (b_{\ell} (v(\tau_t)) + x_{n+1,\ell} )
	< \\ &|\rho_k(\tau_t)|  - \Delta_2({\ell_2^\dagger}) ( |\eta_k(\tau_t)| + x_{n+1,k}).
\end{align}
\end{lemma}
where we defined for a vector $a \in \mathbb{R}^{n+1}$ \\
$$b_{\ell} (a) :=\max \left\{ \sum \limits_{a_i > 0} |a_i| x_{i \ell}, \sum  \limits_{a_i < 0} |a_i| x_{i \ell} \right\}$$
%
%
The Lemma \ref{lemma:lemma_1} essentially states that if the condition in (\ref{eq:lemma_1_eq}) is satisfied, then one can safely ignore the subtree with $\ell$ as the root node, thereby dramatically improving the computational efficiency.
We extended our method to elastic net and call it ENet-SHIM. The details of ENet-SHIM and proof of Lemma \ref{lemma:lemma_1} are provided in the section~\ref{additional_technical_details}.


%
%
%
\subsection{Algorithms}
The algorithms to compute the $\tau$-path and the full-CP set of SHIM are given in Algorithm \ref{algo:tau_path} and Algorithm \ref{algo:full-CP} respectively. The Algorithm \ref{algo:tau_path} returns the transition points ($\mb{T}$), the LASSO solutions ($\mb{B}$) and the active sets ($\mb{A}$) at those transition points  which are subsequently provided to Algorithm \ref{algo:full-CP}. In Algorithm \ref{algo:full-CP}, for each linear piece of the $\tau$-path, i.e. \(\forall \tau \in (\tau_t, \tau_{t+1})\) for any two consecutive kinks $\tau_t$ and $\tau_{t+1}$, we need to first identify the points at which $|w_{n+1}(\tau)| = |w_i(\tau)|, \forall i\in [n]$ (Line 4). Let us denote those points as \(\{\tau_t = u_1, u_2, u_3, \ldots, u_r = \tau_{t+1} \}\) (for simplicity we slightly abused the notation here). Then we need to check that at which of those points the condition stated in (\ref{eqn:CP2}) is satisfied (Line 5, 6, 7) to determine the full-CP set of SHIM. We used the following search range $[y_{min}, y_{max}]$ to construct the $\tau$-path in Algorithm \ref{algo:tau_path}.  
$$[y_{min}, y_{max}] =[y_{(0)} - 0.5 (y_{(n)} - y_{(0)}), y_{(n)} + 0.5 (y_{(n)} - y_{(0)})],$$ 
where $y_{(0)}\leq y_{(1)} \leq \ldots, y_{(n)}$ are the order statistics of the response vector $y$ (see Remark $5$ in \cite{lei2019fast} for details).
%
\begin{algorithm}[h!]
\begin{footnotesize}
\caption{Compute $\tau$-path}
\label{algo:tau_path}
\begin{algorithmic}[1]
\STATE \textbf{Input:} \(Z \in \mb{R}^{(n+1) \times m}, y \in \mb{R}^n, [y_{min}, y_{max}], \lambda \)
\STATE Initialization: 
\STATE\hspace{\algorithmicindent} \( t=1, \tau_1 = y_{min}\), \(\mc{A}_{\tau_1} = \{\ell \in [p] : \beta_\ell(\tau_1) \neq 0\}\), 
\STATE\hspace{\algorithmicindent}\(\mb{T}= \{\tau_1\},  \mb{B}= \{\beta_{\mc{A}_{\tau_1}}(\tau_1)\}, \mb{A}=\{\mc{A}_{\tau_1}\}\)
%
%
\WHILE{\((\tau_t < y_{max})\)}
\STATE Compute \(\Delta_{\ell^\ast} \) using (\ref{eq:step_length}) and Lemma \ref{lemma:lemma_1}
\IF{\(\Delta_{\ell^\ast} = \Delta_1(\ell_{1}^{\ast})\)}
    \STATE \(\mc{A}_{\tau_{t+1}} \leftarrow  \mc{A}_{\tau_t} \setminus \{\ell_1^\ast\} \) \hst \Comment{remove $\ell_1^\ast$ from \(\mc{A}_{\tau_t}\)}
\ENDIF
\IF{$\Delta_{\ell^\ast} = \Delta_2({\ell_2^\ast})$}
    \STATE\(\mc{A}_{\tau_{t+1}} \leftarrow  \mc{A}_{\tau_t} \cup \{\ell_2^\ast\} \) \hst  \Comment{add  \(\ell_2^\ast\) into \(\mathcal{A}_{\tau_t}\)}
\ENDIF
 \STATE Update:
 \STATE\hspace{\algorithmicindent} \( \tau_{t+1} \leftarrow \tau_{t} + \Delta_{\ell^\ast} \)
\STATE\hspace{\algorithmicindent} \(\beta_{\mc{A}_{\tau_t}}(\tau_{t+1}) \leftarrow \beta_{\mc{A}_{\tau_t}}(\tau_t) + \Delta_{\ell^\ast} \nu_{\mc{A}_{\tau_t}}(\tau_t)\)
\STATE\hspace{\algorithmicindent} \(\mb{T}=\mb{T} \cup \{\tau_{t+1}\}\)
\STATE\hspace{\algorithmicindent} \( \mb{B}=\mb{B} \cup \{\beta_{\mc{A}_{\tau_t}}(\tau_{t+1})\} \)
\STATE\hspace{\algorithmicindent} \(\mb{A}=\mb{A} \cup \{\mc{A}_{\tau_{t+1}}\}\)
%
%
\STATE\hspace{\algorithmicindent} t = t +1
\ENDWHILE
\STATE \textbf{Output:} $\mb{T}, \mb{B}, \mb{A}$
\end{algorithmic}
\end{footnotesize}
\end{algorithm}
%
%
%
%
%
\begin{algorithm}[h!]
\begin{footnotesize}
\caption{Compute full-CP}
\label{algo:full-CP}
\begin{algorithmic}[1]
\STATE \textbf{Input:} \(Z \in \mb{R}^{(n+1) \times m}, y \in \mb{R}^n, \alpha \in [0, 1], \mb{T}, \mb{B}, \mb{A}  \)
%
\STATE Initialization: \( t=0, t_{max} = |\mb{T}|, \mc{C} = \{\emptyset\}\)
\WHILE{\((t < t_{max})\)}
%
\STATE $\{u_2, \ldots, u_{r-1} \}=\{u \in (\tau_t, \tau_{t+1}) \text{ such that } \newline \phantom{hhhhhhhhhahhhaaaafdgfd} |w_i(u)| = |w_{n+1}(u)| \forall i \in [n]\}$
\STATE \(\mc{T}_t = \{\tau_t\} \cup \{u_2, u_3, \ldots, u_{r-1} \} \cup \{\tau_{t+1}\} \)
%
%
\STATE \(\mc{H}_t = \{ h \in \{1, \ldots, | \mc{T}_t | \} \, :\,  \pi(\mc{T}_t (h)) \geq \alpha\}\) using $\mb{B}$ and  $\mb{A}$
%
\STATE \(\mc{C} = \mc{C} \cup_{h \in \mc{H}_t} [ \mc{T}_t(h), \mc{T}_t(h + 1)) \)
\STATE t = t +1
\ENDWHILE
\STATE \textbf{Output:} $\mc{C}$
\end{algorithmic}
\end{footnotesize}
\end{algorithm}

%% file: sec4.tex
\section{Results and Discussions}
We evaluated our proposed method using both synthetic and real-world data. For all experiments, we considered a coverage guarantee of $90\%$, that is, \(\alpha = 0.1\). We compared the statistical efficiency of our proposed method (SHIM) with those of other simple (LASSO) and complex models (NN, RF).  For the complex models, we reported the split-CP because it is the only available method for computing the CP for complex models. We also demonstrated the statistical efficiency of full-CP in comparison to that of split-CP both for LASSO and SHIM.  
\subsection{Comparison of statistical efficiencies}
\subsubsection{Synthetic data experiments}
\begin{figure}[t]
   \centering
   \includegraphics[width=\linewidth]{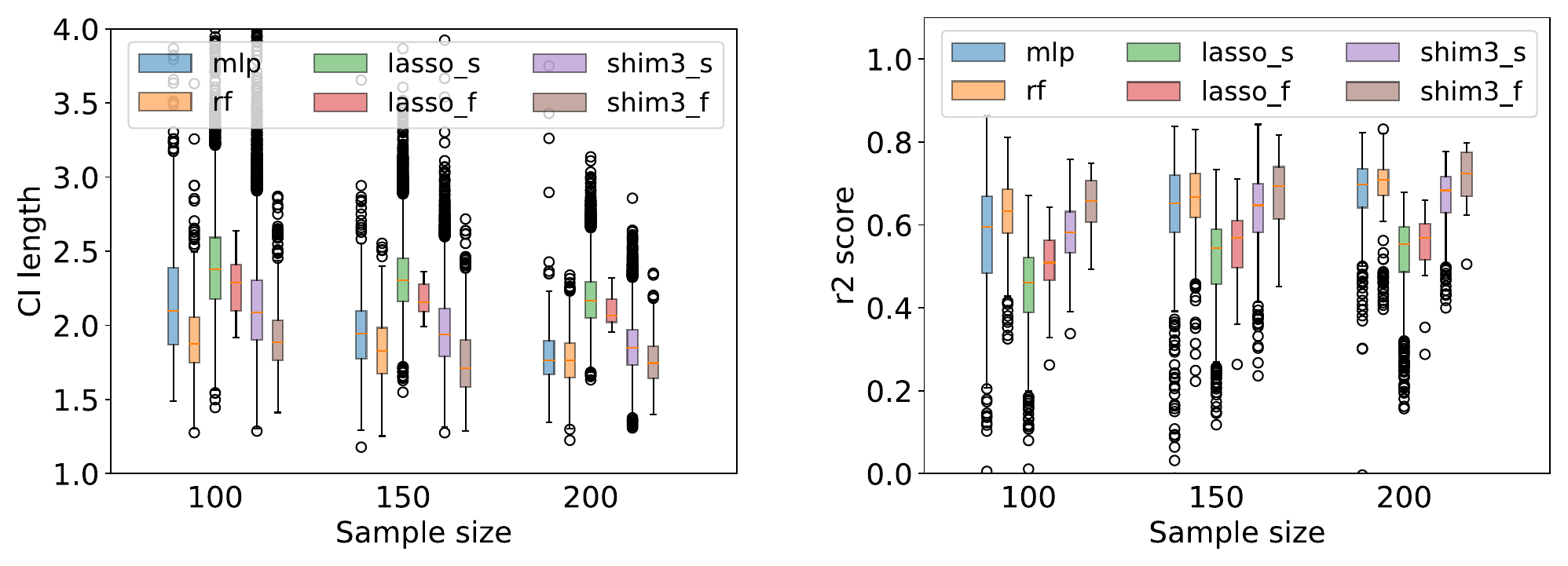}
   \caption{Comparison of the confidence interval lengths (CI length) and r2 scores of the proposed method (shim) with other simple (lasso) and complex models (mlp, rf) using synthetic data for different sample sizes. shim\_3s and shim\_3f respectively represent split-CP and full-CP for a $3^{rd}$ order SHIM.}
   \label{fig:cpl_synthetic}
 \end{figure}

We generated random i.i.d. samples $(z_i, y) \in \{0,1\}^m \times \mb{R} $ in such a way that $100m(1-\zeta)\%$ features of $z_i \in \mb{R}^m$ contain a value of 1 on average.
Here, $\zeta \in [0, 1]$ is the sparsity controlling parameter that controls the sparsity of the design matrix, whereas the sparsity in the model coefficients are controlled by the regularizer $\lambda$.
The pruning effectiveness (Fig. \ref{fig:node_counts} and Table~\ref{table:comp_eff_time_taken}) depends on the sparsity of the design matrix as it exploits the tree's anti-monotonicity property. High dimensional real-world data is generally very sparse and the choice of $\zeta$ in our experiments is just for the demonstration purpose.
The response $y_i \in \mb{R}$ is randomly generated from a normal distribution $N(\mu(x_i), \sigma^2)$. For demonstration purposes, we considered a true model of up to fifth-order interactions, which is defined as $\mu (x_i) = 2.0z_1 +2.0 z_1z_2 + 2.0z_1z_2z_3 + 2.0z_1z_3z_4z_5 + 2.0z_1z_2z_3z_4z_5$, and set $\sigma=1$. The choice of this model is merely for the demonstration purposes, and the proposed method is equally applicable to any chosen model. We used the same true model $\mu(x_i)$ in both low and high dimensional settings (Table~\ref{table:stat_low} \& ~\ref{table:stat_high} and Fig.~\ref{fig:cpl_synthetic}).
%

\begin{figure}[t]
   \centering
   \includegraphics[width=\linewidth]{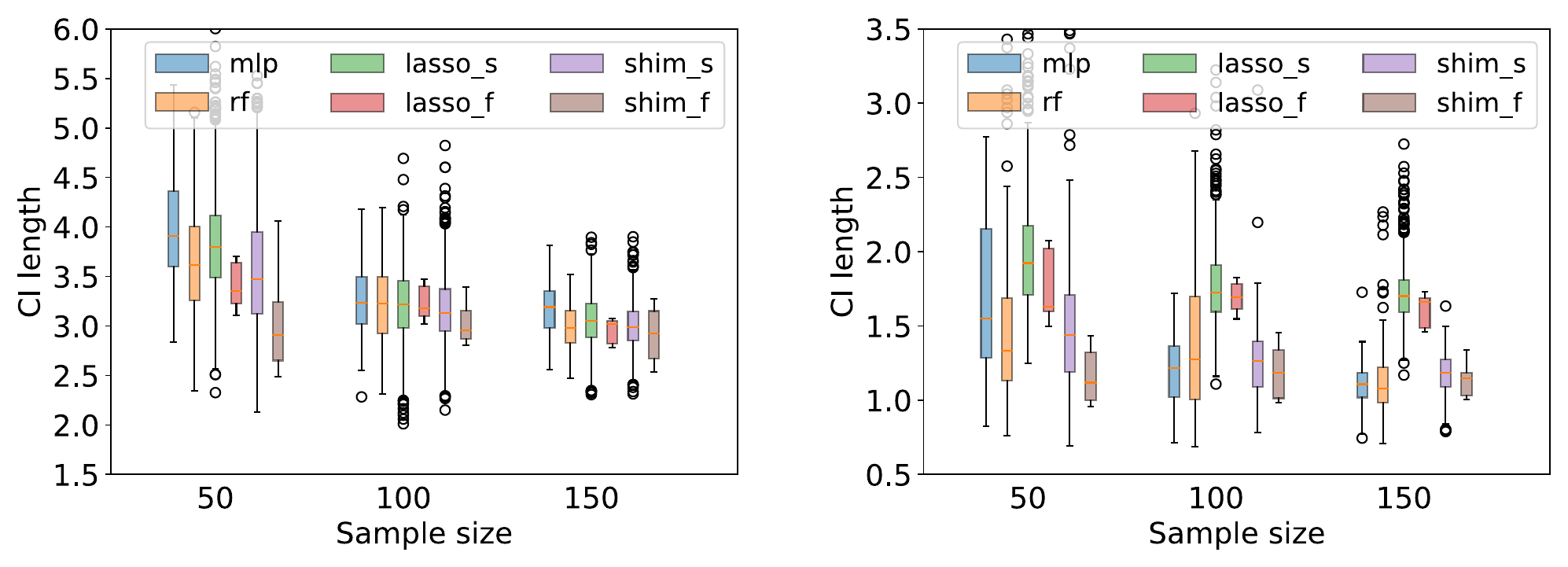}
   \caption{Comparison of the confidence interval lengths (CI length) of the proposed method (shim) with other simple (lasso) and complex models (mlp, rf) using synthetic data for different sample sizes. Left figure shows the results using model-1 (weak signal) and the right figure shows the results using model-2 (strong signal). Here, we did not mention any max order of interaction and the entire search space is used to choose the best SHIM model by the algorithm automatically.}
   \label{fig:syn_low_signal_sparse}
 \end{figure}

\begin{figure}[t]
   \centering
   \includegraphics[width=\linewidth]{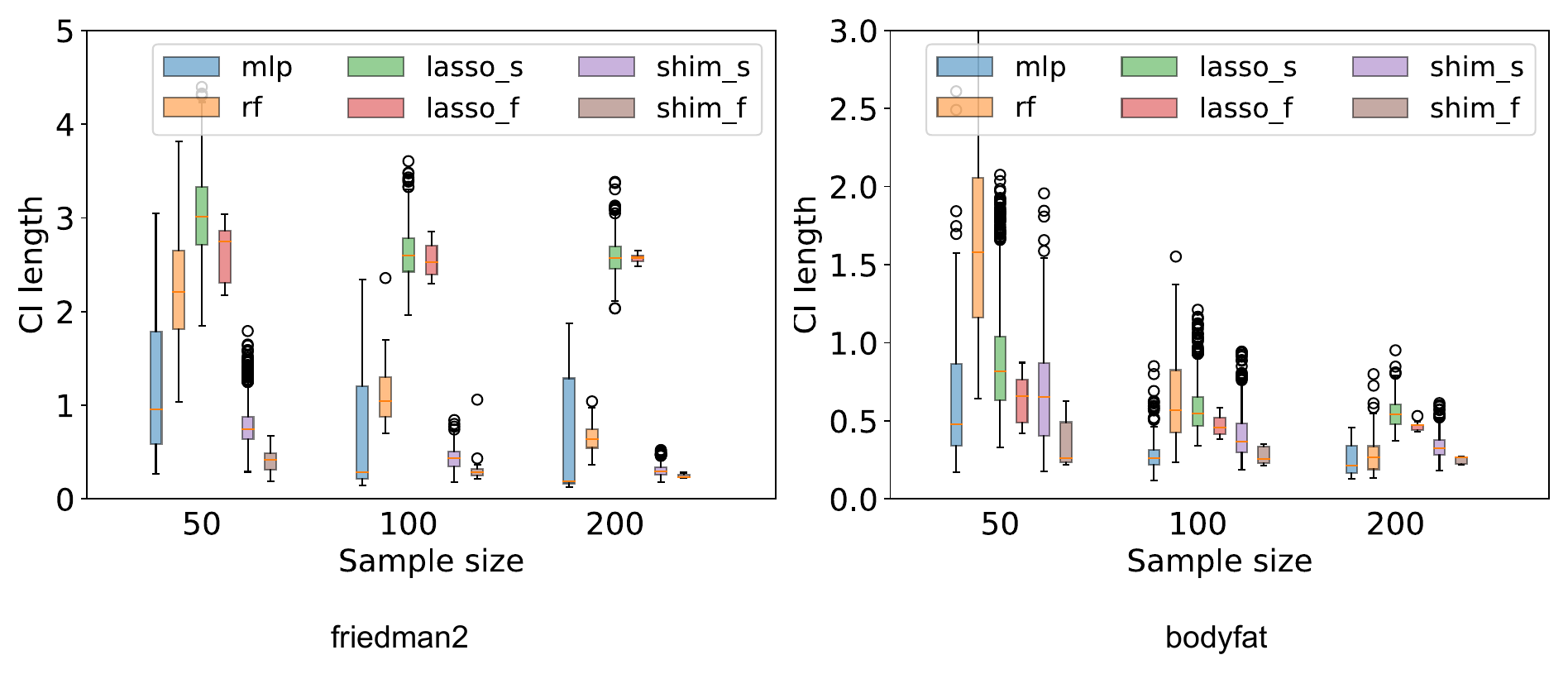}
   \caption{Comparison of the confidence interval lengths (CI length) of the proposed method (shim) with other simple (lasso) and complex models (mlp, rf) using continuous synthetic (friedman2) and real world (bodyfat) data. shim\_s and shim\_f respectively represent split-CP and full-CP for a SHIM. Here, we did not mention any max order of interaction and the entire search space is used to choose the best SHIM model by the algorithm automatically.}
   \label{fig:cpl_continuous}
 \end{figure}

%
In the low-dimensional setting, we considered a dataset with $n=150$ instances and $m=10$ original covariates, whereas for the high-dimensional setting, we considered $n=150$ and $m=100$. We kept the sparsity of the design matrix fixed at $\zeta =0.4$ in both the settings. We varied the order of interaction $d=2, 3, \cdots$, however, almost in all the experiments we found that the model get saturated after $3^{rd}$ order interactions (i.e. the performance of SHIM did not change much for $d\geq 3$). Hence, we reported the results of up to $3^{rd}$ order interactions. 
Note that the max order of interaction $d$ is not needed to be specified beforehand. Our pruning condition takes care of the case even when $d$ is not specified that is when the whole search space is considered (Fig. \ref{fig:node_counts} and Table~\ref{table:comp_eff_time_taken}).
Later, we provided results (Fig.~\ref{fig:syn_low_signal_sparse},~\ref{fig:cpl_hiv} \& ~\ref{fig:cpl_continuous}) where we did not impose any constraints on the max-pat size ($d$).
For both the settings (low and high), we generated a dataset of $n=150$ training instances, each accompanied with $n=50$ test instances. We generated 5 such independent random datasets and repeated the experiments 3 times. Hence, in total we reported the average results of 15 independent datasets i.e. \(3 \times 5 \times 50 = 750 \) test instances. The details of the hyper parameter selection are given in section~\ref{hyperparameter_selection}. We used a multi-layer perceptron (MLP) as a neural network architecture in our experiments.
\begin{table}[t]
\caption{Comparison of the statistical power of the proposed method (shim) with other simple (lasso) and complex models (mlp, rf) using low dimensional synthetic data ($m=10, n=150$). The bracketed values represent the standard deviations.}
\label{table:stat_low}
\centering
\resizebox{\linewidth}{!}{
\begin{tabular}{ |c|c|c|c|c|c|c|c|c| } 
\hline
 &mlp & rf &lasso\_s & lasso\_f &shim\_2s &shim\_2f &shim\_3s &shim\_3f\\
 \hline
 length&\begin{tabular}{c} 1.94\\(0.27) \end{tabular}  & \begin{tabular}{c} 1.84\\ (0.23) \end{tabular}   &\begin{tabular}{c} 2.32\\ (0.23)  \end{tabular}   &\begin{tabular}{c} 2.18\\  (0.10)\end{tabular}   &\begin{tabular}{c} 2.00\\  (0.23) \end{tabular}  &\begin{tabular}{c} 1.82\\ (0.17) \end{tabular}   &\begin{tabular}{c} 1.96 \\ (0.24) \end{tabular} &\begin{tabular}{c} 1.76 \\ (0.25)  \end{tabular}  \\
\hline
  cov&\begin{tabular}{c} 0.90 \\(0.05) \end{tabular} &\begin{tabular}{c} 0.89 \\(0.06) \end{tabular} &\begin{tabular}{c} 0.91 \\(0.04) \end{tabular}  &\begin{tabular}{c} 0.90 \\(0.05) \end{tabular} &\begin{tabular}{c}0.91 \\(0.04) \end{tabular} &\begin{tabular}{c}0.89 \\(0.05) \end{tabular}  &\begin{tabular}{c}0.90 \\(0.04) \end{tabular} &\begin{tabular}{c}0.88 \\(0.04) \end{tabular} \\
\hline
 r2&\begin{tabular}{c}0.61 \\(0.18)\end{tabular} &\begin{tabular}{c} 0.66 \\(0.09)\end{tabular} &\begin{tabular}{c}0.51 \\(0.12)\end{tabular}  &\begin{tabular}{c}0.54 \\(0.11)\end{tabular} &\begin{tabular}{c}0.63 \\(0.10)\end{tabular} &\begin{tabular}{c}0.67 \\(0.10)\end{tabular}  &\begin{tabular}{c}0.63 \\(0.09)\end{tabular} &\begin{tabular}{c}0.67 \\(0.09) \end{tabular}\\
 \hline
   \end{tabular}}
\end{table}
\begin{table}[t]
\caption{Comparison of the statistical power of the proposed method (shim) with other simple (lasso) and complex models (mlp, rf) using high dimensional synthetic data ($m=100, n=150$). The bracketed values represent the standard deviations.}
\label{table:stat_high}
\centering
\resizebox{\linewidth}{!}{
\begin{tabular}{ |c|c|c|c|c|c|c|c|c| } 
\hline
 &mlp & rf &lasso\_s
      & lasso\_f & shim\_2s &shim\_2f  &shim\_3s &shim\_3f\\
 \hline
length&\begin{tabular}{c}3.39 \\(0.51)\end{tabular} &\begin{tabular}{c} 1.93 \\(0.32)\end{tabular} &\begin{tabular}{c} 2.45 \\(0.40)\end{tabular}  &\begin{tabular}{c} 2.26 \\(0.19)\end{tabular} &\begin{tabular}{c} 2.20 \\(0.36)\end{tabular} &\begin{tabular}{c}1.92 \\(0.26)\end{tabular}  &\begin{tabular}{c}2.27 \\(0.38)\end{tabular} &\begin{tabular}{c}1.89 \\(0.30)\end{tabular} \\
  \hline
  cov&\begin{tabular}{c}0.91 \\(0.05)\end{tabular} &\begin{tabular}{c} 0.90 \\(0.06)\end{tabular} &\begin{tabular}{c} 0.90 \\(0.06)\end{tabular}  &\begin{tabular}{c} 0.92 \\(0.05)\end{tabular} &\begin{tabular}{c}0.90 \\(0.05)\end{tabular} &\begin{tabular}{c}0.91 \\(0.04)\end{tabular}  &\begin{tabular}{c}0.90 \\(0.06)\end{tabular} &\begin{tabular}{c}0.90 \\(0.04)\end{tabular} \\
  \hline
 r2&\begin{tabular}{c}-0.01 \\(0.16)\end{tabular} &\begin{tabular}{c} 0.66 \\(0.11)\end{tabular} &\begin{tabular}{c}0.45 \\(0.09)\end{tabular}  &\begin{tabular}{c}0.53 \\(0.07)\end{tabular} &\begin{tabular}{c}0.54 \\(0.13)\end{tabular} &\begin{tabular}{c}0.65 \\(0.09)\end{tabular}  &\begin{tabular}{c}0.52 \\(0.12)\end{tabular} &\begin{tabular}{c}0.66 \\(0.08)\end{tabular} \\
 \hline
   \end{tabular}}
\end{table}
\vspace{0.2cm}
From Table \ref{table:stat_low} and Table \ref{table:stat_high} it can be observed that both in low- and high-dimensional settings, all the methods produced perfect (or nearly perfect) coverage i.e cov=$0.90$. The statistical efficiency of individual methods are compared using the length of confidence interval. A statistically efficient model is expected to produce a shorter confidence interval length. Comparing the length of confidence intervals, one can observe that the statistical efficiency of SHIM increases as we increase the order of interactions. 
A $3^{rd}$ order SHIM produced the shortest average confidence interval lengths both in low- and high-dimensional settings. 
To compare the fitting power of individual methods we reported the $R$-squared (r2) scores. It can be observed that the r2 scores of a $3^{rd}$ order SHIM are comparable to the best performing complex model (RF) both in low- and high-dimensional settings. We also demonstrated the comparison of the confidence interval lengths (CI length) and $R$-squared (r2) scores of the proposed method (SHIM) with other simple (LASSO) and complex models (MLP, RF) for three different sample sizes, $n \in \{100, 150, 200\}, m=10$ (Fig. \ref{fig:cpl_synthetic}). It can be observed that the full-CP of $3^{rd}$ order SHIM (shim3\_f ) produced the best results (shortest avg. length, highest avg. r2 score) in all the cases. It can also be observed that the performance of data splitting (in mlp, rf, lasso\_s, shim3\_s) is worse compared to that of full-CP. The split-CP tends to produce a longer confidence interval and smaller r2 score, and suffer from high variance due to the smaller data size as well as the additional randomness considered in data splitting.
\vspace{0.2cm}

To further compare the performance of the proposed method we considered highly sparse data ($\zeta=0.6$) in two different experimental settings.
(1) \textbf{model-1}: We generated a model with weak signals that is we considered a true model of up to third-order interactions, which is defined as $\mu (x_i) = 1.0z_1 +1.0 z_1z_2 + 1.0z_1z_2z_3$ and (2) \textbf{model-2}: We generated a model with strong signals that is we considered a true model of up to third-order interactions, which is defined as $\mu (x_i) = 5.0z_1 +5.0 z_1z_2 + 5.0z_1z_2z_3$  . In both the settings we set $\sigma=1$. The choice of these models is merely for demonstration purposes, and the proposed method is equally applicable to any chosen model. 
For both the settings (weak and strong), we generated a dataset of $n\in \{50, 100, 150\}$ training instances, each accompanied with $n=10$ test instances. Here we considered a covariate size of $m=5$. We generated 3 such independent random datasets. Hence, in total we reported the average results of  \(3 \times 10 = 30 \) test instances. 
%
For split-CP, we repeated the experiments 30 times to highlight the effect of randomness in the confidence set generation. For SHIM, we did not mention any max-pat size of $d$, i.e. the entire search space is considered for exploration (tree generation) and the proposed tree pruning condition takes care of it to improve the efficiency of the search. The best shim model is automatically chosen by the algorithm corresponding to the best hyper parameter $\lambda$ (see section~\ref{hyperparameter_selection} for details of hyper parameter selection).
The results are shown in Fig. \ref{fig:syn_low_signal_sparse} where the left figure corresponds to the model-1 (weak signal)  and the right figure corresponds to the model-2 (strong signal). In both the settings, the full-CP methods (lasso\_f, shim\_f) produced more compact confidence sets irrespective of sample sizes.
It can be observed that in case of highly sparse data, the shim model produces better results, more specifically when the sample size is small and the signal is weak.\looseness=-1

\subsubsection{Real-world data experiments.} 
\begin{figure}[t]
   \centering
   \includegraphics[width=\linewidth]{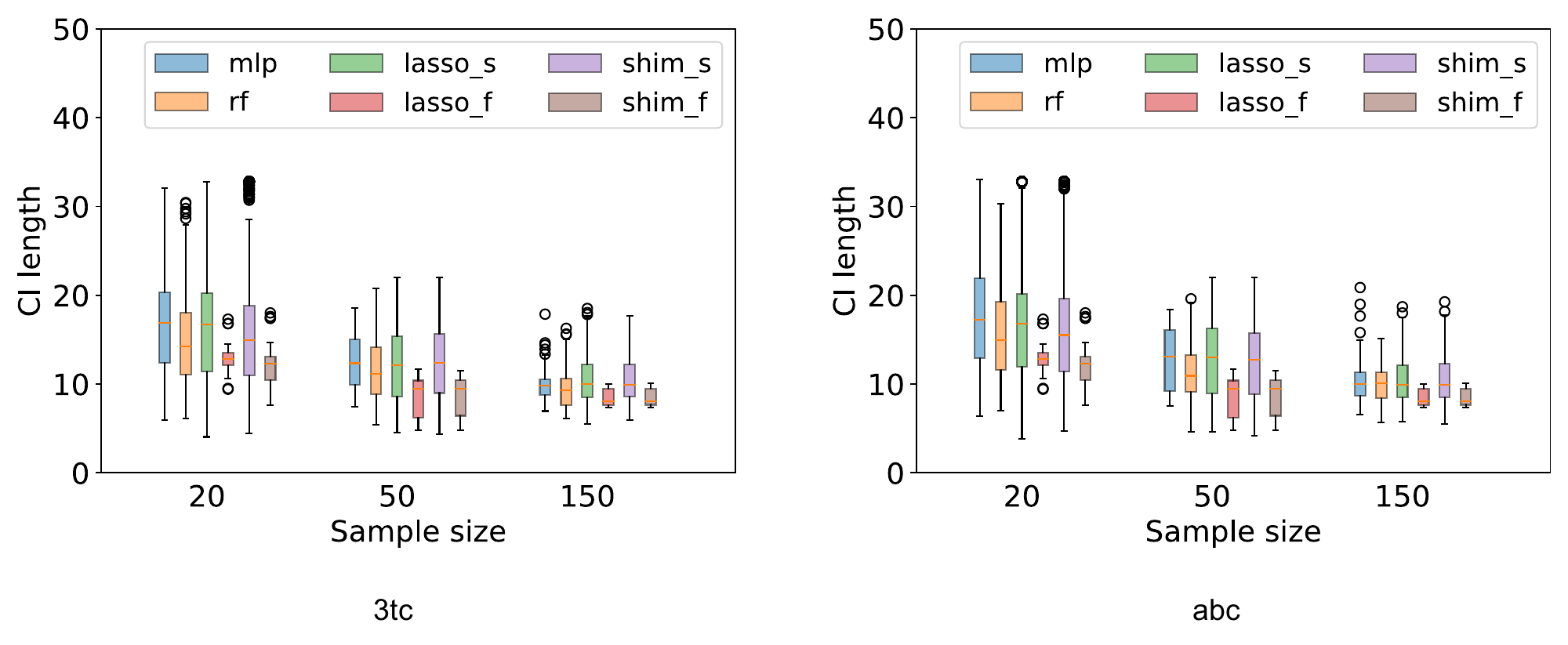}
   \caption{Comparison of the confidence interval lengths (CI length) of the proposed method (shim) with other simple (lasso) and complex models (mlp, rf) using two hiv drug resistance data ('3tc' and 'abc') for different sample sizes. shim\_s and shim\_f respectively represent split-CP and full-CP for a SHIM. Here, we did not mention any max order of interaction and the entire search space is used to choose the best SHIM model by the algorithm automatically.}
   \label{fig:cpl_hiv}
 \end{figure}
%
%
%
%

\textbf{HIV Data.} We applied our method on real world HIV drug resistance dataset. The HIV-1 sequence data was obtained from the Stanford HIV Drug Resistance Database \cite{rhee2003human,rhee2006genotypic}. We applied our method on two NRTI drugs, Lamivudine (3TC) and Abacavir (ABC). The results are shown in Fig.~\ref{fig:cpl_hiv}. Here, we considered top ten mutations for the result generations. In HIV data set most of the columns contain zeros, hence, we sorted them based on the number of 1s present in each column and selected the top ten columns. We subsampled the data to generate independent training sets for three different training sizes, $n \in \{20, 50, 150\}$, each accompanied with a separate test set consisting of 10 instances. We repeated the experiment three times and in total we reported the results of $3 \times 10 = 30$ test instances. 

\subsubsection{Experiments with continuous data.}
To demonstrate the efficacy of our method using continuous data we considered both synthetic and real world continuous data. \textbf{Synthetic data:} we considered the standard Friedman2 data generated by the scikit-learn package. This data contains four independent features randomly distributed on specific intervals (for details please see scikit-learn package). 
\textbf{Real data:} we considered the bodyfat data obtained from the LIBSVM repository for regression data analysis.  This data set consists of 252 instances and 14 numerical features (for details please see LIBSVM repository). For both synthetic and real world data, we rescaled all the features values in the range of $0$ and $1$ and conducted all the experiments with rescaled features values.
 Like before, we subsampled the data to generate independent training sets for three different training sizes, $n \in \{50, 100, 200\}$, each accompanied with a separate test set consisting of 10 instances. We repeated the experiment three times and in total we reported the results of $3 \times 10 = 30$ test instances. The results are shown in Fig. \ref{fig:cpl_continuous}. \looseness=-1


%
%
%
%
%
\subsection{Comparison of computational efficiencies.}\label{sec_comp_eff}
\begin{table*}[t]
\caption{Computation time (in sec) with and without pruning using two different $\lambda$ values ($\lambda=1, 10$) for three different sparsity levels ($\zeta=0.4, 0.7, 0.9$). All computation times were measured on an Intel(R) Xeon(R) Gold 6130 CPU @ 2.10GHz.}
\label{table:comp_eff_time_taken}
\centering
\resizebox{\textwidth}{!}{
\begin{tabular}{ |c|c|c|c|c|c|c|c|c|c|c|c|c|c| } 
\hline
 \multirow{4}{*}{$d$} & \multirow{4}{*}{\shortstack{Search space \\ (\# nodes)}} &  \multicolumn{6}{|c|}{$\lambda=1$} & \multicolumn{6}{|c|}{$\lambda=10$}\\
 \cline{3-14}
& & \multicolumn{3}{|c|}{With pruning} &\multicolumn{3}{|c|}{Without pruning} &\multicolumn{3}{|c|}{With pruning} &\multicolumn{3}{|c|}{Without pruning}\\
\cline{3-14}
& &$\zeta=0.4$ &$\zeta=0.7$ &$\zeta=0.9$ &$\zeta=0.4$ &$\zeta=0.7$ &$\zeta=0.9$ &$\zeta=0.4$ &$\zeta=0.7$ &$\zeta=0.9$ &$\zeta=0.4$ &$\zeta=0.7$ &$\zeta=0.9$\\
 \hline
 \hline
  2 & 465 &0.09  &0.10   &0.03  &0.10  &0.08  &0.07   &0.08  &0.08 &0.03  &0.07  &0.07 &0.12 \\
  \hline
  3 & 4525 &0.84  &0.67   &0.03  &0.82  &0.79 &0.64 &0.56  &0.36 &0.03  &0.63  &0.65 &0.85  \\
  \hline
  4 & 31930 &4.49   &1.51   &0.03  &5.91  &5.32 &4.73 &2.14  &0.90 &0.03  &4.12  &4.42 &3.87  \\
  \hline
  5 & 174436 &12.34    &2.74   &0.03  &30.19  &28.76    &23.31  &5.11  &1.38 &0.03  &26.74  &22.70 &25.42  \\
  \hline
  10 & 53009101 &112.17    &3.83   &0.03  &$>$ 1 day  &$>$ 1 day &$>$ 1 day &49.88  &2.13 &0.03  &$>$ 1 day  &6891.44 &6861.16  \\
  \hline
 15 & 614429671 &126.04   &3.39   &0.03  &$>$ 1 day  &$>$ 1 day &$>$ 1 day &56.22   &2.00 &0.03  &$>$ 1 day  &$>$ 1 day & $>$ 1 day  \\
 \hline
 20 & 1050777736  &126.86    &3.51   &0.03   &$>$ 1 day  &$>$ 1 day &$>$ 1 day &55.33  &2.13 &0.03   &$>$ 1 day  &$>$ 1 day & $>$ 1 day \\
  \hline
 25 &1073709892  &130.28   &3.62   &0.03   &$>$ 1 day  &$>$ 1 day &$>$ 1 day &55.21   &2.02 &0.03   &$>$ 1 day  &$>$ 1 day &$>$ 1 day  \\
 \hline
\end{tabular}}
\end{table*}
\begin{figure}[h!]
   \centering
      \includegraphics[width=\linewidth]{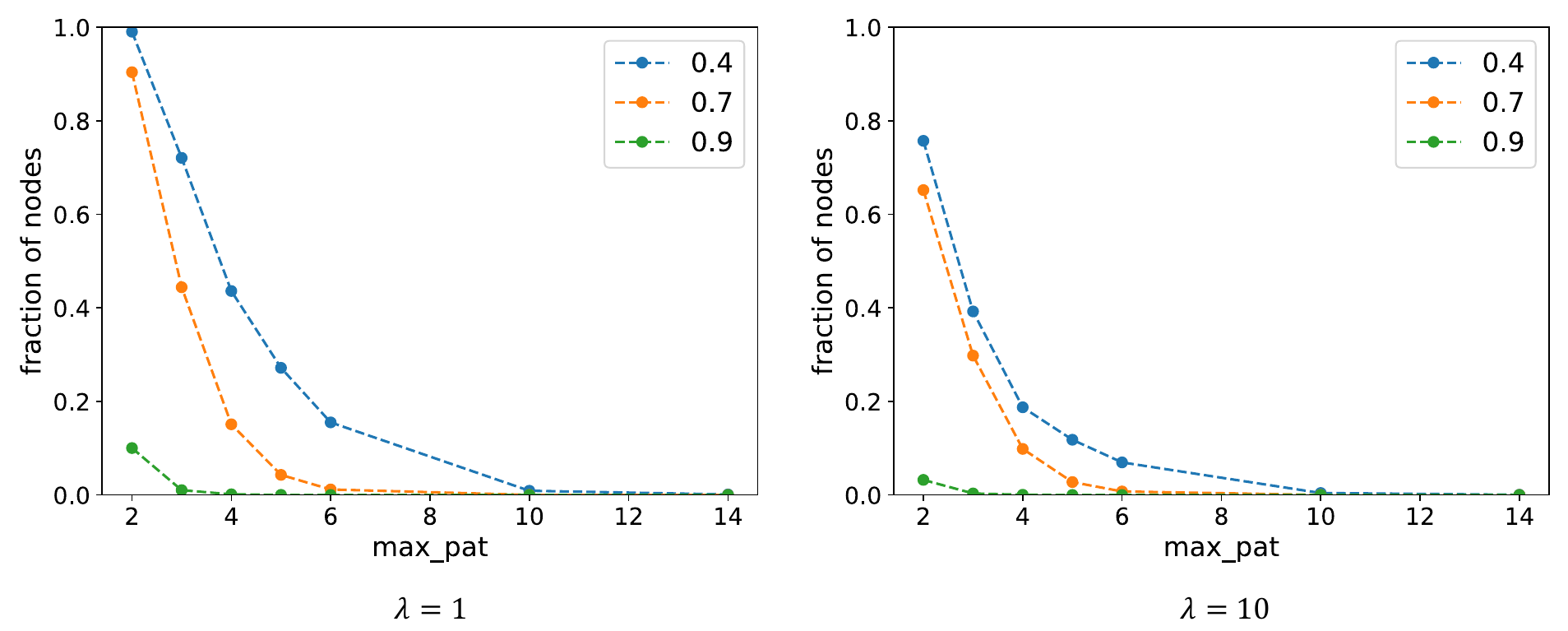}
   \caption{Variation of node counts (``fraction of node counts") for different \emph{max\_pat} ($d$) sizes. Here, the fraction of nodes for a specific $d$, represents the number of nodes traversed divided by the total number of possible combinations of interaction terms using the ``max-pat" size of $d$. The results are shown for two $\lambda$ values ($\lambda = 1, 10$) and for three different sparsity levels ($0.4, 0.7, 0.9$). The pruning is more effective as the data is more sparse. }
   \label{fig:node_counts}
 \end{figure}
\begin{table}[t]
\caption{Number of homotopy transition points (\# kinks) along the $\tau$-path of SHIM using two different $\lambda$ values ($\lambda=1, 10$) for three different sparsity levels ($\zeta=0.4, 0.7, 0.9$) for different value of ``max\_pat" ($d$) sizes.}
\label{table:comp_eff_kinks}
\centering
\resizebox{\linewidth}{!}{
\begin{tabular}{ |c|c|c|c|c|c|c|c| } 
\hline
 \multirow{2}{*}{$d$} & \multirow{2}{*}{\shortstack{Search space \\ (\# nodes)}} &  \multicolumn{3}{|c|}{$\lambda=1$} & \multicolumn{3}{|c|}{$\lambda=10$}\\
 \cline{3-8}
 & &$\zeta=0.4$ &$\zeta=0.7$ &$\zeta=0.9$  &$\zeta=0.4$ &$\zeta=0.7$ &$\zeta=0.9$\\
 \cline{3-8}
 \hline
 \hline
  2 & 465 &200  &134   &7  &9  &7  &2    \\
  \hline
  3 & 4525 &317  &129   &7  &9  &7  &2    \\
  \hline
  4 & 31930 &314   &129   &7  &9  &7  &2    \\
  \hline
  5 & 174436 &312    &129   &7  &9  &7  &2    \\
  \hline
  10 & 53009101 &312    &129   &7  &9  &7  &2   \\
  \hline
 15 & 614429671 &312   &129   &7  &9  &7  &2    \\
 \hline
 20 & 1050777736  &312    &129   &7   &9  &7  &2   \\
  \hline
 25 &1073709892  &312   &129   &7   &9  &7  &2   \\
 \hline
\end{tabular}
}
\end{table}
To demonstrate the computational efficiency of the proposed pruning strategy for the $\tau$-path, we generated a synthetic dataset of $n=100$ and $m=30$ for three different sparsity levels of the design matrix ($\zeta=0.4, 0.7, 0.9$)  using the same $5^{th}$ order model as used to demonstrate the statistical efficiency in Table~\ref{table:stat_low} \& ~\ref{table:stat_high} and Fig.~\ref{fig:cpl_synthetic}. 
We compared both the fraction of nodes traversed (Fig. \ref{fig:node_counts}) and the time taken (Table \ref{table:comp_eff_time_taken}) against a different maximum interaction order $d$  for three different sparsity levels ($\zeta=0.4, 0.7, 0.9$) using two different $\lambda$ values ($\lambda=1, 10$). 
%
%
It can be observed that the pruning is more effective at the deeper nodes of the tree and saturates after a certain depth of the tree.
This is evident as the sparsity of the data increases at the deeper nodes, and the pruning exploits the anti-monotonicity of high-order interaction terms constructed as tree of patterns.

In the case of the homotopy method without pruning, we stopped the execution of the program if the $\tau$-path was not finished in one day.
From Table ~\ref{table:comp_eff_time_taken}, it can be observed that without the tree pruning, the construction of the $\tau$-path is not practical as we progress to the deeper nodes of the tree because of the generation of an exponential number of high-order interaction terms.
The maximum time taken by the $\tau$-path with pruning was approximately 130 s, for a ``max-pat" size of 25, i.e. for 1073709892 number of nodes at $\lambda=1, \zeta=0.4$.
In our numerical experiments, the number of kink in the $\tau$-path appears to be polynomial in the number of features (Table \ref{table:comp_eff_kinks}). 
The worst-case complexity of $\tau$-path is exponential (see section~\ref{proposed_method}). However, fortunately, it has been well-recognized that this worst-case rarely happens in practice \cite{li2018well,le2021parametric} and, this is also evident from our experimental results (Table \ref{table:comp_eff_kinks}). \looseness=-1
%
%
%

%% file: conclusion.tex
\section{Conclusion}
In this paper we proposed an algorithm to efficiently compute the full-CP set of SHIM. Through numerical experiments, we demonstrated that SHIM  is statistically superior than the vanilla lasso and comparable to other complex models (NN, RF). The confidence interval along with a point estimation is better than a point estimation alone. The computation of a full-CP set for other complex models (NN, RF) remains an open question. SHIM is interpretable and it generates the decision function as a weighted combination of \emph{decision sets} (if-then rules). The homotopy-mining based exact full-CP of SHIM is better than the split-CP of SHIM. The efficient pruning strategy makes the computation of exact full-CP set tractable for SHIM.

%% file: appendix-A.tex
\section{Additional Technical Details}\label{additional_technical_details}
\subsection{Proof of Lemma \ref{lemma:lemma_1}}
%
Before proving the Lemma \ref{lemma:lemma_1}, we introduce the following two propositions:
\begin{proposition}\label{prop:1} 
Let's define for a vector $a \in \mathbb{R}^{n+1}$
\begin{align*}
 	 b_{\ell}(a) &= \max \left\{ \sum_{a_i < 0} \lvert a_i \lvert x_{i\ell}, \sum_{a_i > 0} \lvert a_i \lvert x_{i\ell} \right\},
\end{align*}
then if we expand $\rho_\ell (\tau_t)$ and $\eta_\ell (\tau_t)$ separately for positive and negative values of $w_i(\tau_t)$ and $v_i(\tau_t), \forall i \in [n+1]$ respectively,  we can write
\begin{align*}
	|\rho_\ell (\tau_t)| & \leq b_{\ell}(w(\tau_t)), \\ 
	|\eta_\ell (\tau_t)| & \leq b_{\ell}(v(\tau_t)),
\end{align*}
\end{proposition}
\vspace{0.5cm}
%
%
\begin{proof}[Proof of Proposition \ref{prop:1}]
We have 
\begin{align*}
	|x_\ell^\top a| &=  \left | \sum 
	\limits_{i=1}^{n+1} a_i x_{i \ell} \right |  \\ 
	&=  \left | \sum 
	\limits_{a_i > 0} |a_i| x_{i \ell} - \sum  \limits_{a_i < 0} |a_i| x_{i \ell}\right | \\ 
	&\leq  \max \left \{ \sum 
	\limits_{a_i > 0} | a_i| x_{i \ell}, \sum  \limits_{a_i < 0} |a_i| x_{i \ell} \right \} \\ &=: b_{\ell}(a).
\end{align*}
%
\end{proof}
Note that $\rho_{\ell} = x_{\ell}^{\top}w(\tau_t)  \text{ and }
    \eta_{\ell} = x_{\ell}^{\top} v(\tau_t)$, and here we used a generic vector $a$ in place of $w(\tau_t)$ and $v(\tau_t)$ to keep the proof simple.
\begin{proposition}\label{prop:2}
By using the tree anti-monotonicity property \ie, $x_{i\ell} \geq x_{i\ell^\prime}, \hst \forall \ell^\prime \supset \ell, \forall i \in [n+1]$ as defined in the definition of tree (section~\ref{tree_pruning}), we have 
\begin{align*}
	b_{\ell}(a) & \geq b_{\ell^\prime}(a),  
\end{align*}
\end{proposition}
%
\begin{proof}[Proof of Proposition \ref{prop:2}]
From the definition of tree we have $x_{i\ell}\geq x_{i\ell^\prime}, \forall \ell^{\prime} \supset \ell, \forall i \in [n+1]$. Hence, we can write
\begin{align*}
	b_{\ell}(a) &= \max \left\{ \sum_{a_i < 0} \lvert a_i \lvert x_{i\ell}, \sum_{a_i > 0} \lvert a_i \lvert x_{i\ell} \right\} \\ 
	&\geq \max \left\{ \sum_{a_i < 0} \lvert a_i \lvert x_{i\ell^\prime}, \sum_{a_i > 0} \lvert a_i \lvert x_{i\ell^\prime} \right\}\\
	&=: b_{\ell^\prime}(a).
\end{align*}
\end{proof}
Now, for any node $\ell^\prime$ such that $\ell^\prime \supset \ell$, we can write (\ref{eqn:inclusion_condition2}) as follows.
\begin{align}\label{eqn:inclusion_condition3}
\nonumber
    \big\lvert \rho_{\ell^\prime}(\tau_t) -   &\Delta_2(\ell^\prime) (\eta_{\ell^\prime}(\tau_t) -  x_{n+1,\ell^\prime})   \lvert \\  &= \big\lvert \rho_{k}(\tau_t) -  \Delta_2(\ell^\prime) (\eta_{k}(\tau_t) - x_{n+1,k} )\big\lvert \end{align}
%
%

Here, we used the fact that $$\beta_{\mc{A}_{\tau_t}}(\tau_{t+1})=\beta_{\mc{A}_{\tau_t}}(\tau_t) + \Delta_2(\ell^\prime)\nu_{\mc{A}_{\tau_t}}(\tau_t).$$
The right hand side (r.h.s.) of (\ref{eqn:inclusion_condition3}) has a lower bound i.e.
\begin{align*}
    \lvert \rho_{k}(\tau_t) -   &\Delta_2(\ell^\prime) (\eta_{k}(\tau_t) - x_{n+1,k})  \lvert \\  & \geq 
    \lvert \rho_{k}(\tau_t) \rvert  -   \Delta_2(\ell^\prime) (\lvert \eta_{k}(\tau_t) \rvert  +   \lvert x_{n+1,k} \rvert), \end{align*}
and the left hand side (l.h.s.) of (\ref{eqn:inclusion_condition3}) has an upper bound i.e.
\begin{align*}
    \lvert \rho_{\ell^\prime}(\tau_t) -  &\Delta_2(\ell^\prime) (\eta_{\ell^\prime}(\tau_t) - x_{n+1,\ell^\prime} )  \lvert \\ &\leq 
    \lvert \rho_{\ell^\prime}(\tau_t) \rvert  + \Delta_2(\ell^\prime) (\lvert \eta_{\ell^\prime}(\tau_t) \rvert  +  \lvert x_{n+1,\ell^\prime} \rvert). \end{align*}
The above two bounds are derived by considering the fact that for any $a \in \mb{R}, b \in \mb{R},  c \in \mb{R}$ and $d \in \mb{R}>0$ we can write the following:
\begin{align*}
&|a-d(b-c)| \leq |a| + d (|b|+|c|) \hst \text{ and } \\  &|a-d(b-c)| \geq |a| - d (|b|+ |c|).
\end{align*}
Therefore, for (\ref{eqn:inclusion_condition3}) to have a solution the following condition needs to be satisfied.
\begin{align}\label{eqn:inclusion_condition4}
\nonumber
    \lvert \rho_{\ell^\prime}(\tau_t) \rvert + &\Delta_2(\ell^\prime) (\lvert \eta_{\ell^\prime}(\tau_t) \rvert +  \lvert x_{n+1,\ell^\prime}   \lvert ) \\ &\geq \lvert \rho_{k}(\tau_t) \rvert  -   \Delta_2(\ell^\prime) (\lvert \eta_{k}(\tau_t) \rvert  +  \lvert x_{n+1,k} \rvert).
\end{align}
Hence, (\ref{eqn:inclusion_condition3}) will not have any solution if the following condition (\ref{cond:pruning_tau}) is satisfied.
%
%
%
\begin{align}\label{cond:pruning_tau}
\nonumber
\lvert \rho_{\ell^\prime}(\tau_t) \lvert + &\Delta_2(\ell^\prime) (\lvert \eta_{\ell^\prime}(\tau_t) \lvert + x_{n+1,\ell^\prime}) \\ &< \lvert \rho_k (\tau_t) \lvert   - \Delta_2(\ell^\prime) ( \lvert\eta_k(\tau_t) \lvert + x_{n+1,k}).
\end{align}
%
%
%
%
Now, using Proposition \ref{prop:1} we can further write (\ref{eq:lemma_3_eq}) which implies (\ref{cond:pruning_tau}).
\begin{align} \label{eq:lemma_3_eq}
	\nonumber b_{\ell^\prime}(w(\tau_t)) + &\Delta_2(\ell^\prime) (b_{\ell^\prime}( v(\tau_t)) + x_{n+1,\ell^\prime})  \\
	&< |\rho_k(\tau_t)|  - \Delta_2(\ell^\prime) (|\eta_k(\tau_t)| + x_{n+1,k}).
\end{align}
\begin{proof}[Proof of Lemma \ref{lemma:lemma_1}]
We now prove Lemma \ref{lemma:lemma_1} by contradiction, that is we assume that at any node $\ell$, the condition (\ref{eq:lemma_1_eq}) stated in Lemma~\ref{lemma:lemma_1} holds, and there exists one \( \ell^{\prime} \supset \ell:  \Delta_2({\ell^{\prime}}) < \Delta_2({\ell_2^\dagger}) \); then show that this is a contradiction.\looseness=-1
%
%
%
%
\begin{align*}
    \therefore \quad &|\rho_k(\tau_t)| - \Delta_2({\ell^{\prime}}) (|\eta_k(\tau_t)| + x_{n+1,k})\\
    &> |\rho_k(\tau_t)| - \Delta_2({\ell_2^\dagger}) (|\eta_k(\tau_t)| + x_{n+1,k}),\\ & \hspace{4cm} \because \Delta_2({\ell^{\prime}}) < \Delta_2({\ell_2^\dagger})\\
    &> b_{\ell}(w(\tau_t)) + \Delta_2({\ell_2^\dagger}) (b_{\ell}( v(\tau_t)) + x_{n+1,\ell}),\\ &\hspace{4cm} \text{using } (\ref{eq:lemma_1_eq})\\
    &> b_{\ell^\prime}(w(\tau_t)) + \Delta_2({\ell_2^\dagger}) (b_{\ell^\prime} (v(\tau_t)) + x_{n+1,\ell^\prime}),\\ & \hspace{4cm}\text{(Proposition \ref{prop:2}),}\\
    &> b_{\ell^\prime}(w(\tau_t)) + \Delta_2({\ell^\prime}) (b_{\ell^\prime} (v(\tau_t)) + x_{n+1,\ell^\prime}),\\
    & \hspace{4cm} \because \Delta_2({\ell^{\prime}}) < \Delta_2({\ell_2^\dagger}).
\end{align*}
Therefore, we got
\begin{align*}
  |\rho_k(\tau_t)| - &\Delta_2({\ell^{\prime}}) (|\eta_k(\tau_t)|  + x_{n+1,k})\\
  &> b_{\ell^\prime}(w(\tau_t)) + \Delta_2({\ell^\prime}) (b_{\ell^\prime} (v(\tau_t)) + x_{n+1,\ell^\prime}) \\
  &\implies \ell^{\prime} \text{ is infeasible},\\
  &\hspace{1.5cm}\text{(using (\ref{eq:lemma_3_eq}))}\\
  &\implies \Delta_2({\ell^{\prime}}) \geq \Delta_2({\ell_2^\dagger}).
\end{align*}
\end{proof}
This completes the proof of Lemma \ref{lemma:lemma_1}.
Hence, if the pruning condition in Lemma \ref{lemma:lemma_1} holds, then we do not need to search the sub-tree with $\ell$ as the root node, and hence increasing the efficiency of the search procedure.
%
%

%% file: appendix-B.tex
%
\subsection{Extension for Elastic Net (ENet-SHIM)}\label{el-SHIM}
A common problem of the LASSO is that if the data has correlated features, then the LASSO picks only one of them and ignores the rest, which leads to instability. To solve this problem \cite{zou2005regularization} proposed the Elastic Net. This feature correlation problem is very much evident in the SHIM-type problem, and hence we extended our framework for the Elastic Net. We solve the following optimization problem to extend our framework for the Elastic Net:
\begin{equation}\label{obj:primal_elnet}
    \beta(\tau) \in \argmin_{\beta \in \bbR^p} \frac{1}{2}\norm{y(\tau) - X\beta}_2^2 + \frac{1}{2}\alpha \norm{\beta}_2^2  + \lambda \norm{\beta}_1.
\end{equation}
The elastic net optimization problem can actually be formulated as a LASSO optimization problem using augmented data.
If we consider an augmented data defined as \(\Tilde{X} = \begin{pmatrix} X\\ \sqrt{\alpha} I_p \end{pmatrix}  \in \mb{R}^{(n+1+p) \times p}\) and \(\Tilde{y} (\tau) = \begin{pmatrix}y (\tau) \\ \0 \end{pmatrix} \in \mb{R}^{n+1+p}\), where $I_p \in \mb{R}^{p \times p}$ is an identity matrix and $\0 \in \mb{R}^p$ is a zero vector, then solving the elastic net optimization problem (\ref{obj:primal_elnet}) for a fixed $\lambda$, is equivalent to solving the following problem.
\begin{equation*}
    \beta(\tau) \in \argmin_{\beta \in \bbR^p} \frac{1}{2}\norm{\Tilde{y} (\tau) - \Tilde{X}\beta}_2^2 + \lambda \norm{\beta}_1.
\end{equation*}
\textbf{Step size ($\tau$-path):}
If we consider two real values $\tau_t$ and $\tau_{t+1}$ ( $\tau_{t+1}>\tau_t$) at which the active set does not change, and their signs also remain the same, then we can write 
$$\beta_{\mathcal{A}_{\tau_{t}}}(\tau_{t+1}) - \beta_{\mathcal{A}_{\tau_t}}(\tau_t) = \nu_{\mathcal{A}_{\tau_t}}(\tau_t)(\tau_{t+1} - \tau_t),$$
%
%
where
$$\nu_{\mc{A}_{\tau_t}}(\tau_t) = (X^{\top}_{\mc{A}_{\tau_t}}X_{\mc{A}_{\tau_t}} + \alpha I_{|\mathcal{A}_{\tau_t}|} )^{-1}x_{n+1;\mc{A}_{\tau_t}}.$$

Note that here the only change compared to the vanilla SHIM is the addition of an \(\alpha I_{|\mathcal{A}_{\tau_t}|}\) term to the expression of \(\nu_{\mathcal{A}_{\tau_t}}(\tau_t)\). Now, one can also derive a similar expression of the step size of inclusion and deletion as done for the vanilla SHIM $\tau$-path (\ref{eq:step_length}), considering the updated expression of \(\nu_{\mathcal{A}_{\tau_t}}(\tau_t)\).
\subsubsection{Tree pruning ($\tau$-path)}
The no solution condition (\ref{cond:pruning_tau}) with the augmented data (\(\Tilde{X}, \Tilde{y}\)) can be written as follows:
\begin{align}\label{cond:pruning_elnet1}
\nonumber
\lvert \Tilde{\rho}_{\ell^\prime}(\tau_t) \lvert + &\Delta_2(\ell^\prime) (\lvert \Tilde{\eta}_{\ell^\prime}(\tau_t) \lvert + x_{n+1,\ell^\prime}) \\
&< \lvert \Tilde{\rho}_k (\tau_t) \lvert   - \Delta_2(\ell^\prime) (\lvert\Tilde{\eta}_k(\tau_t) \lvert + x_{n+1,k}),
\end{align}
where
\begin{align*}
\Tilde{\rho}_{\ell^\prime} (\tau_t) &= \Tilde{x}_{\ell^\prime}^{\top}\Tilde{w}(\tau_t) \text{ and} \\
\Tilde{\eta}_{\ell^\prime} (\tau_t) &=\Tilde{x}_{\ell^\prime}^{\top} \Tilde{v}(\tau_t), \forall \ell^\prime \in \mathcal{A}^c_{\tau_t}, \\ \Tilde{\rho}_k (\tau_t) &= \Tilde{x}_{k}^{\top}\Tilde{w}(\tau_t) \text{ and}  \\ \Tilde{\eta}_k (\tau_t) &= \Tilde{x}_k^{\top} \Tilde{v}(\tau_t), \forall k \in \mathcal{A}_{\tau_t}, \\
\Tilde{w}(\tau_t) &= \Tilde{y} (\tau_t) - \Tilde{X}_{\mc{A}_{\tau_t}}\beta_{\mc{A}_{\tau_t}}(\tau_t)  \in \mathbb{R}^{n+1+p}, \\
\Tilde{v}(\tau_t) &= \Tilde{X}\nu_{\tau_t}  \in \mathbb{R}^{n+1+p}.
\end{align*}
Now one can show that the tree pruning condition for the $\tau$-path of ENet-SHIM can be defined as stated in Lemma \ref{lemma:lemma_2}.
\begin{lemma}\label{lemma:lemma_2}
%
For any given node $\ell$, if $\Delta_2(\ell_2^\dagger)$ is the current minimum step size, that is,
$$
\ell_{2}^{\dagger} = \argmin_{j \in \{1, 2, \ldots, \ell\}\cap \mc{A}_{\tau_t}^c} \Delta_2(j),
$$
%
then
$\forall \ell^\prime \supset \ell, \Delta_2({\ell^\prime}) \geq \Delta_2({\ell_2^\dagger})$ if 
\begin{align} \label{eq:lemma_2_eq}\nonumber
	 b_{\ell}(w(\tau_t)) + &\Delta_2({\ell_2^\dagger}) (b_{\ell}(v(\tau_t)) + x_{n+1,\ell})\\
	&< |\Bar{\rho}_k(\tau_t)|  - \Delta_2({\ell_2^\dagger}) ( |\Bar{\eta}_k(\tau_t)| + x_{n+1,k}),
\end{align}
\end{lemma}
where
\begin{align*}
\Bar{\rho}_k(\tau_t) &= \sum_{i=1}^n w_i(\tau_t) x_{ik} - \alpha \beta_k, \\
\Bar{\eta}_k(\tau_t) &= \sum_{i=1}^n v_i(\tau_t) x_{ik} + \alpha \nu_k.
\end{align*}
%
%
\vspace{0.1cm}

Although, theoretically the standard elastic net can be solved as a LASSO optimization problem by considering an augmented dataset as explained in section~\ref{el-SHIM}, this data augmentation can be prohibitively expensive in case of ENet-SHIM due to the combinatorial effects of interaction terms. 
Actually, by expanding the expression of $\Tilde{\rho}_{\ell^\prime} (\tau_t)$, $\Tilde{\rho}_{k} (\tau_t)$, $\Tilde{\eta}_{\ell^\prime} (\tau_t)$ and $\Tilde{\eta}_{k} (\tau_t)$ separately for the original data and the augmented part of the augmented data, one can show that most of the terms will disappear, and one just need to consider additional terms $-\alpha \beta_k(\tau_t)$ and $\alpha\nu_k (\tau_t)$ while evaluating the expression of $\Tilde{\rho}_k (\tau_t)$ and $\Tilde{\eta}_{k} (\tau_t)$ respectively. However, $\tilde{\rho}_{\ell^\prime} (\tau_t)$ and $\tilde{\eta}_{\ell^\prime} (\tau_t)$  will remain the same as in the original data, i.e., $\tilde{\rho}_{\ell^\prime} (\tau_t)$=$\rho_{\ell^\prime} (\tau_t)$ and $\tilde{\eta}_{\ell^\prime} (\tau_t)$=$\eta_{\ell^\prime} (\tau_t)$. The formal proof of Lemma \ref{lemma:lemma_2} is given below.
%
\begin{proof}[Proof of Lemma \ref{lemma:lemma_2}]
\textnormal{Let's consider} 
$$\Tilde{w}(\tau_t) = \Tilde{y} (\tau_t) - \Tilde{X}_{\mc{A}_{\tau_t}}\beta_{\mc{A}_{\tau_t}}(\tau_t)  \in \mathbb{R}^{n+1+p}$$
\textnormal{and}
$$w(\tau_t) = y(\tau_t) - \bar{X}_{\mc{A}_{\tau_t}}\beta_{\mc{A}_{\tau_t}}(\tau_t)  \in \mathbb{R}^{n+1},$$
\textnormal{where} \(p = |\mathcal{A}_{\tau_t}| + |\mathcal{A}^c_{\tau_t}| \), \textnormal{then we can write}
\begin{equation}\label{eqn:aug_w}
    \Tilde{w}_i(\tau_t) = \begin{cases}
     w_i(\tau_t)  &\textnormal{if} \quad  i \leq n+1, \\
    -\sqrt{\alpha}\beta_{\ell^\prime}(\tau_t)  &\textnormal{if} \quad n+1 < i \leq n + 1+ |\mathcal{A}_{\tau_t}|,
    \\  & \qquad \qquad \qquad \qquad \phantom{hh} \forall \ell^\prime \in \mathcal{A}_{\tau_t}, \\
    0  &\textnormal{if} \quad n+1 + |\mathcal{A}_{\tau_t}| < i \leq n + 1 + p.
    \end{cases}
\end{equation}
\textnormal{Similarly considering \(\Tilde{v}(\tau_t) = \Tilde{X}\nu(\tau_t)  \in \mathbb{R}^{n+1+p}\) and \(v(\tau_t) = \bar{X}\nu(\tau_t)  \in \mathbb{R}^{n+1}\), we can write}
\begin{equation}\label{eqn:aug_v}
    \Tilde{v}_i(\tau_t) = \begin{cases}
     v_i(\tau_t) &\textnormal{if} \quad i \leq n +1,\\
    \sqrt{\alpha}\nu_{\ell^\prime}(\tau_t)  &\textnormal{if} \quad n+1 < i \leq n +1 + |\mathcal{A}_{\tau_t}|,
    \hspace{0.5cm} \\ &\qquad \qquad \qquad \qquad \phantom{hh} \forall \ell^\prime \in \mathcal{A}_{\tau_t},\\
    0 &\textnormal{if} \quad n+1 + |\mathcal{A}_{\tau_t}| < i \leq n +1 + p,
    \end{cases}
\end{equation}
\textnormal{and, considering \(\Tilde{X} \in \mathbb{R}^{(n+1+p)\times p}\) and \(X \in \mathbb{R}^{(n+1) \times p}\) we can write}
\begin{equation}\label{eqn:aug_X}
    \Tilde{x}_{i\ell^\prime} = \begin{cases}
    x_{i\ell^\prime} &\textnormal{if} \quad i \leq n+1, \\
    \sqrt{\alpha}  &\textnormal{if} \quad i > n+1 \enspace \textnormal{and} \enspace (i-n-1)=\ell^\prime,\\
    0 &\textnormal{otherwise}.
    \end{cases}
\end{equation}
\textnormal{Therefore, \(\forall \ell^\prime \in \mathbb{R}^p\)  we can write}
\begin{align*}
    \Tilde{\rho}_{\ell^\prime} (\tau_t) &= \Tilde{x}_{\ell^\prime}^{\top} \Tilde{w}(\tau_t)\\ &= \sum_{i=1}^{n+1+p} \Tilde{w}_i(\tau_t) \Tilde{x}_{i\ell^\prime} \\
    &=\sum_{i=1}^{n+1} \Tilde{w}_i(\tau_t) \Tilde{x}_{i\ell^\prime} \hst + \sum_{\substack{i=n+2}}^{n+1+|\mathcal{A}_{\tau_t}|} \Tilde{w}_i(\tau_t) \Tilde{x}_{i\ell^\prime} \hst + \sum_{i=n+2+|\mathcal{A}_{\tau_t}|}^{n+p} \Tilde{w}_i(\tau_t) \Tilde{x}_{i\ell^\prime}.
\end{align*}
\textnormal{Now, using (\ref{eqn:aug_w}) and (\ref{eqn:aug_X}) the second and the third quantity in the above expression can be written as follows:}
\begin{equation*}
\sum_{\substack{i=n+2}}^{n+1+|\mathcal{A}_{\tau_t}|} \Tilde{w}_i (\tau_t) \Tilde{x}_{i\ell^\prime} =\begin{cases}
    (-\sqrt{\alpha}\beta_{\ell^\prime}(\tau_t))(\sqrt{\alpha}), \\ \hspace{1.5cm} \textnormal{if} \enspace (i-n-1) = \ell^\prime,\\
    0 \hspace{0.5cm} \textnormal{otherwise},
    \end{cases}
\end{equation*}
\textnormal{and,}
\begin{equation*}
   \sum_{i=n+2+|\mathcal{A}_{\tau_t}|}^{n+p} \Tilde{w}_i(\tau_t) \Tilde{x}_{i\ell^\prime} = 0. 
\end{equation*}
\textnormal{Therefore,}
\begin{align*}
\Tilde{\rho}_{\ell^\prime} (\tau_t) &=
    \Tilde{x}_{\ell^\prime}^{\top} \Tilde{w}(\tau_t)\\ &=\sum_{i=1}^{n+1} w_i(\tau_t) x_{i\ell^\prime} \\
    &= \rho_{\ell^\prime}(\tau_t), \enspace \forall \ell^\prime \in \mathcal{A}^c_{\tau_t} \enspace \textnormal{and,} 
\end{align*}
\begin{align*}
\Tilde{\rho}_k (\tau_t) &=
   \Tilde{x}_k^{\top} \Tilde{w}(\tau_t)\\
   &=\sum_{i=1}^{n+1} w_i(\tau_t) x_{ik} - \alpha \beta_k (\tau_t)\\
   &= \rho_k(\tau_t) - \alpha \beta_k (\tau_t), \enspace \forall k \in \mathcal{A}_{\tau_t}\\
   &=: \bar{\rho}_k(\tau_t) .
\end{align*}
\textnormal{Similarly, using (\ref{eqn:aug_v}) and (\ref{eqn:aug_X}) we can write}
\begin{align*}
    \tilde{\eta}_{\ell^\prime} (\tau_t)
    &=\Tilde{x}_{\ell^\prime}^{\top} \Tilde{v}(\tau_t)\\ &= \sum_{i=1}^{n+1} v_i (\tau_t) x_{i\ell^\prime} \\
    &= \eta_{\ell^\prime}(\tau_t), \enspace \forall \ell^\prime \in \mathcal{A}_{\tau_t}^c, \hspace{0.5cm} \textnormal{and,}
\end{align*}
\begin{align*}
\tilde{\eta}_k (\tau_t) &=
  \Tilde{x}_k^{\top} \Tilde{v}(\tau_t)\\
  &=\sum_{i=1}^{n+1} v_i(\tau_t) x_{ik} + \alpha \nu_k (\tau_t) \\
  &=  \eta_k(\tau_t) + \alpha \nu_k (\tau_t) , \enspace \forall k \in \mathcal{A}_{\tau_t}\\
  &= \bar{\eta}_k(\tau_t).
\end{align*}
\end{proof}
%
%
Therefore, we can write (\ref{cond:pruning_elnet1}) as follows.
\begin{align}\label{cond:pruning_elnet3}
\nonumber
\lvert \rho_{\ell^\prime}(\tau_t) \lvert + &\Delta_2(\ell^\prime) (\lvert \eta_{\ell^\prime}(\tau_t) \lvert + x_{n+1,\ell^\prime}) \\ &< \lvert \bar{\rho}_k (\tau_t)  \lvert   - \Delta_2(\ell^\prime) (\lvert \bar{\eta}_k(\tau_t) \lvert + x_{n+1,k}),
\end{align}

Now, using Proposition\ref{prop:1}, we can simplify (\ref{cond:pruning_elnet3}) as follows.
\begin{align}\label{pruning_elnet3}
\nonumber
 b_{\ell^\prime,w(\tau_t)} + &\Delta_2(\ell^\prime)  (b_{\ell^\prime,v(\tau_t)}  + x_{n+1,\ell^\prime}) \\ &<  |\Bar{\rho}_k(\tau_t)| - \Delta_2(\ell^\prime) (|\Bar{\eta}_k(\tau_t)| + x_{n+1,k}),
\end{align}
Now similar to the Lemma \ref{lemma:lemma_1} one can formally prove the Lemma \ref{lemma:lemma_2} using (\ref{pruning_elnet3}) and Proposition \ref{prop:2}.
%

%% file: appendix-C.tex
%
\subsection{Hyper parameter selection}\label{hyperparameter_selection}
The hyper parameter selection for all the methods (e.g. $\lambda$ in LASSO, SHIM; hidden layer's sizes in MLP etc.) are done based on $5$-fold cross validation using a separate set of 15 independent samples. For MLP, the activations are chosen from \{identity, relu, logistic, tanh\} and the hidden layer's sizes are chosen from all possible combinations of \{50, 100, 150\} nodes, considering both 2-hidden layers and 3-hidden layers architectures. The most frequent activation and architecture chosen based on separate set of 15 independent samples are considered to report the average results. Similarly for RF, the number of estimators (n\_estimators) and the min samples in the leaf of a tree (min\_samples\_leaf) are chosen from \{50, 100, 200\} and \{0.1, 0.05, 0.01\} respectively. The median $\lambda$ value based on the separate set of 15 independent samples is considered to report the results of LASSO and SHIM. For the $\lambda$ selection we considered the range of $\lambda$ defined in $[\lambda_{max}/2, 0)$. For, MLP and RF, we used the standard \emph{scikit-learn} implementation.

%% file: acknowledgement.tex
This work was partially supported by MEXT KAKENHI (20H00601, 16H06538), JST CREST (JPMJCR21D3), JST Moonshot R\&D (JPMJMS2033-05), NEDO (JPNP18002, JPNP20006) and RIKEN Center for Advanced Intelligence Project.